\documentclass{article}

\usepackage{microtype}
\usepackage{graphicx}
\usepackage{subcaption}
\usepackage{booktabs} % for professional tables

\usepackage{hyperref}

\usepackage[accepted]{icml2023}

\usepackage{amsmath}
\usepackage{amssymb}
\usepackage{mathtools}
\usepackage{amsthm}

\usepackage[capitalize,noabbrev]{cleveref}

\theoremstyle{plain}

\theoremstyle{definition}

\theoremstyle{remark}

\usepackage[textsize=tiny]{todonotes}

\icmltitlerunning{Submission and Formatting Instructions for ICML 2023}

\begin{document}

\twocolumn[
\icmltitle{ConvGenVisMo: Evaluation of Conversational Generative Vision Models}



\icmlsetsymbol{equal}{*}

\begin{icmlauthorlist}
%\icmlauthor{Narjes Nikzad Khasmakhi}{equal,yyy}
\icmlauthor{Narjes Nikzad Khasmakhi}{yyy}
\icmlauthor{Meysam Asgari-Chenaghlu}{comp1}
\icmlauthor{Nabiha Asghar}{comp2}
\icmlauthor{Philipp Schaer}{yyy}
\icmlauthor{Dietlind Zühlke}{yyy}

%\icmlauthor{Firstname8 Lastname8}{sch}
%\icmlauthor{Firstname8 Lastname8}{yyy,comp}
\end{icmlauthorlist}

\icmlaffiliation{yyy}{TH Köln - University of Applied Sciences, Cologne, Germany}
\icmlaffiliation{comp1}{Ultimate, Berlin, Germany}
\icmlaffiliation{comp2}{Microsoft, Redmond WA, USA (This work is not sponsored by or done at Microsoft)}

\icmlcorrespondingauthor{Narjes Nikzad Khasmakhi}{narjes.nikzad\_khasmakhi@th-koeln.de}

\icmlkeywords{Conversational models \and Generative models \and Dialogue Systems \and Evaluation metrics \and Large Language Models \and ChatGPT \and DreamStudio}

\vskip 0.3in
]



\printAffiliationsAndNotice{}  % leave blank if no need to mention equal contribution

\begin{abstract}
  Conversational generative vision models (CGVMs) like Visual ChatGPT \cite{wu2023visual} have recently emerged from the synthesis of computer vision and natural language processing techniques. These models enable more natural and interactive communication between humans and machines, because they can understand verbal inputs from users and generate responses in natural language along with visual outputs. To make informed decisions about the usage and deployment of these models, it is important to analyze their performance through a suitable evaluation framework on realistic datasets. %However, to the best of our knowledge, no such datasets or evaluation frameworks exist today. 
  In this paper, we present \textit{ConvGenVisMo}, a framework %to bridge these two gaps in 
  for the novel task of evaluating CGVMs. \textit{ConvGenVisMo} introduces a new benchmark evaluation dataset for this task, and also provides a suite of existing and new automated evaluation metrics to evaluate the outputs. All \textit{ConvGenVisMo} assets, including the dataset and the evaluation code, will be made available publicly on \href{https://github.com/nabihach/ConvGenVisMo}{Github}\footnote{\texttt{https://github.com/nabihach/ConvGenVisMo}}.
\end{abstract}

\section{Introduction}
\label{Introduction}

Conversational models are AI-powered systems that use natural language processing (NLP) and machine learning algorithms to carry out human-like conversations with users. The state of the art has rapidly advanced in this area with the advent of large language models (LLMs) like GPT-4~\cite{openai2023gpt4}, LaMDA~\cite{thoppilan2022lamda}, PaLM~\cite{chowdhery2022palm} and Llama~\cite{touvron2023llama}, as well as training methods like Reinforcement Learning from Human Feedback (RLHF)~\cite{christiano2017deep}. Two prominent examples of LLM-based conversational models are OpenAI's ChatGPT\footnote{\url{https://chat.openai.com/}} and Google's Bard\footnote{\url{https://blog.google/technology/ai/bard-google-ai-search-updates/}}. They demonstrate excellent command over language syntax and semantics, can carry out fluent multi-hop conversations with humans without losing long-term context, and are able to perform few-shot question answering with surprisingly high accuracy. 
%They also have high versatility in linguistic capabilities, e.g., joke or trivia generation, writing entire blogs and creative stories from scratch, and tailoring responses to each individual prompt given by the user. Perhaps equally important is their high efficiency and scalability, allowing them to deal with large volumes of queries and respond in real time. 

An important question about LLMs is,
 can they turn conversations with humans into actions? 
 For instance, state of the art text-to-image models like Stable Diffusion~\cite{DBLP:journals/corr/abs-2112-10752} and DALL-E~\cite{DBLP:journals/corr/abs-2102-12092} can understand textual prompts to generate highly complex images, but this is a single-round process; they are not able to generate and/or refine images iteratively over the course of an entire conversation. Training a multi-modal conversational model is one feasible solution for turning conversations with humans into actions. However, this approach is prohibitive because it requires a lot of training data to tune the large number of parameters needed in such multi-modal models. The next best alternative is to synthesize a trained conversation model with a trained vision model to obtain a \textbf{conversational generative vision model (CGVM)}. A user can converse with a CGVM's language component, which outputs a summary of the conversation and passes it to the CGVM's visual component, which in turn outputs a realistic image relevant to the conversation. 
 Visual ChatGPT \cite{wu2023visual} is the first effort to synthesize LLMs with image generation models. 
 
 It is reasonable to expect CGVMs to become ubiquitous in the near future, which leads to the need for an appropriate performance evaluation framework for them. However, due to the novelty of the task CGVMs perform, there is no benchmark dataset available to assess their output quality. Moreover, it is not clear how existing image quality assessment metrics should be used to evaluate these models.

\begin{figure*}[h]
    \centering
    \includegraphics[width=0.7 \textwidth]{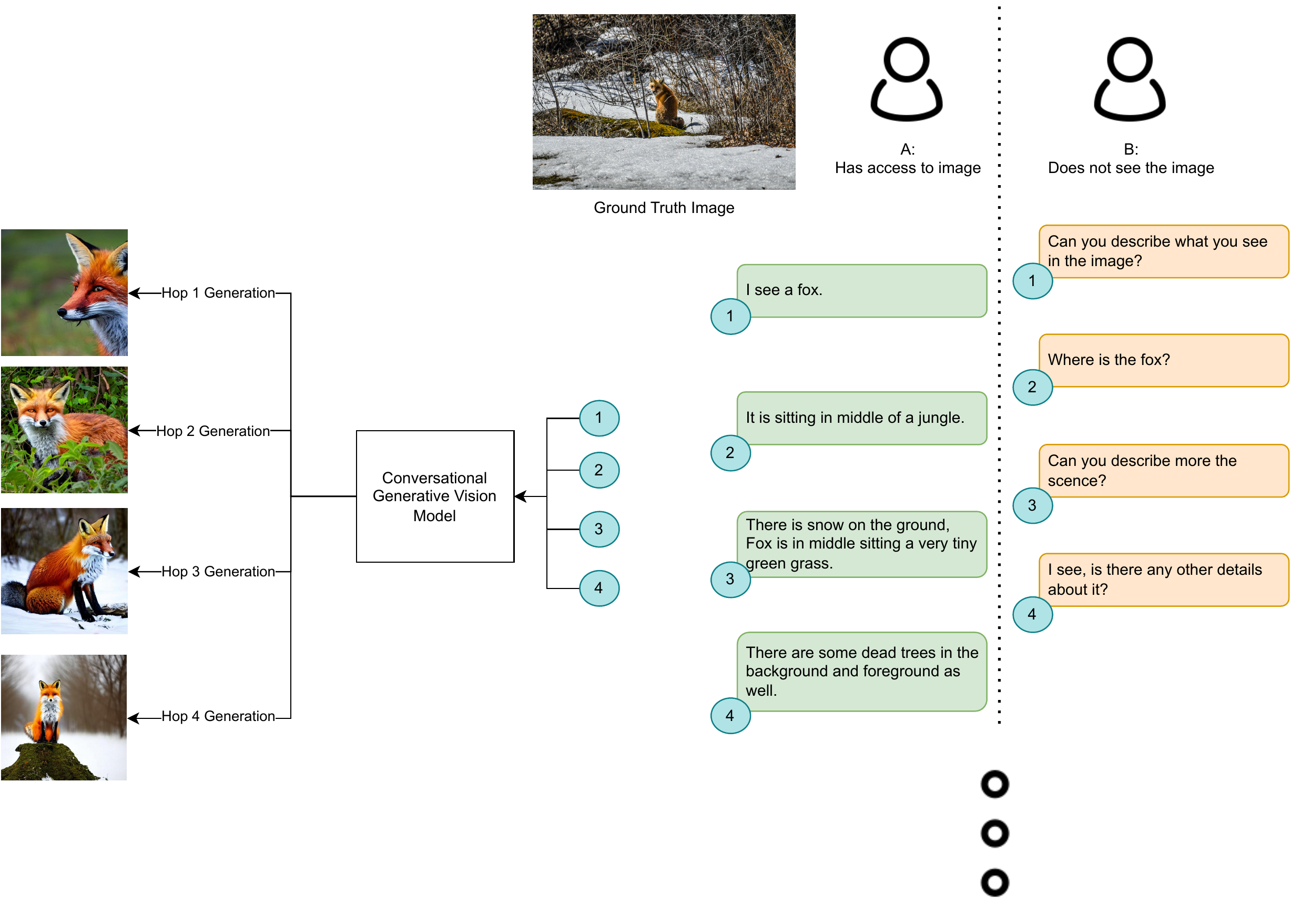}
    \caption{Overview of CGVM evaluation process. Two users converse about a ground truth image. Then the CGVM summarizes conversations upto each hop, and generates an image corresponding to each summary. Finally, the generated images are compared with the ground truth image using the evaluation metrics.}
    \label{fig:example}
\end{figure*}

In this study, we develop two ideas to bridge the aforementioned gaps in CGVM evaluation. 

First, we curate a new benchmark dataset for CGVM evaluation. It contains 180 manually collected and carefully labelled image-conversation pairs\footnote{We will increase the number of samples in future work.}. Each sample in the dataset mimics the scenario where a human is conversationally instructing a model to generate and refine an image. Concretely, each sample contains an image, and a conversation between two real users about that image. Only one user has access to the image, and is explaining its contents to the second user who does not have access to the image. The second user listens to the first user's description of the image and may ask questions about it to get a better understanding of it. %Each conversation is carried out at three different language difficulty levels: basic, intermediate, and advanced.\footnote{This categorization allows us to assess the LLMs' capability to understand different language complexities.}. 
The dataset provides six categories of images: product, nature, human, animal, cartoon, and paintings. These categories allow broad coverage of conversational topics. Note that we curate this new dataset of images from scratch and purposefully do not use existing image datasets like MS COCO \cite{lin2014microsoft}, PASCAL VOC \cite{Everingham10}, or ImageNet \cite{deng2009imagenet}, because pre-trained language and vision models may have already seen them and may perform unrealistically well on them. 

Second, we collect two groups of new and existing evaluation metrics that are appropriate for CGVM evaluation. The first group consists of semantic and computation metrics that assess the quality of the CGVM generated images with and without reference images. The second group uses a novel Element Presence Score to assess element-based overlap between the generated and ground-truth images.

Figure~\ref{fig:example} shows an overview of the CGVM evaluation process.

Thus, our main contribution in this work is an evaluation framework for CGVMs, called \textit{ConvGenVisMo}:
\begin{itemize}
  \item \textit{ConvGenVisMo} introduces the CGVM evaluation task, and the first benchmark dataset for it. The dataset consists of unseen and carefully curated image-conversation pairs spanning several image categories.% and multiple language complexity levels.
  \item \textit{ConvGenVisMo} provides a suite of several existing and new evaluation metrics to assess the performance of CGVMs. 
  \item We demonstrate \textit{ConvGenVisMo}'s usage and results on a CGVM that is obtained from the synthesis of ChatGPT and Dreamstudio\footnote{\url{https://beta.dreamstudio.ai/generate}}. % with more models slated for inclusion in future work. 
  %\item We also evaluate some conversational generative models using the proposed measures and dataset.
\end{itemize}

\section{\textit{ConvGenVisMo}: Task and Dataset}
\label{second_section}

The task at hand is to evaluate the quality of a conversational generative vision model (CGVM). By definition, a CGVM is a model that carries out natural language conversations with humans and produces visual outputs if needed. For example, it can respond to questions about an image it generates, and can handle image editing instructions to further refine the image. 

For the CGVM evaluation task described above, we introduce the \textit{ConvGenVisMo} benchmark dataset which contains 180 samples, each containing an image and a multi-hop conversation between two humans about that image. The image is revealed to only one user, whose goal is to describe it to the other user through a back and forth conversation. The person who does not have access to the image commences the conversation. This individual is referred to as ``Joe'' in the dataset. The second individual, named ``Jill'', provides descriptions of the image, pointing out the presence or absence of specific elements and explaining their significance. When Joe does not understand certain aspects of the image, he seeks clarification. For the sake of simplicity, we only use text-based conversations in the dataset. Inclusion of speech-based conversations is set aside as future work.
Note that this dataset can also be used for evaluating visual question answering models, because it contains contextual questions and answers with respect to an image.

We define the dataset mathematically as follows. Each image $Y_i$, where $1 \leq i \leq 180$, is associated with a conversation $C_i$. Each conversation $C_i$ is a time-series given by

\begin{equation}\label{eq:conversation}
    C_i = \{(M_{Joe}, M_{Jill})_t : t \in \mathbb{N}\}_i,
\end{equation}

where the tuple $(M_{Joe}, M_{Jill})_t$ is the $t$'th conversation hop consisting of the message sent by Joe to Jill and the response sent by Jill to Joe at step $t$. Since all images are inaccessible to Joe, every conversation starts with him asking questions to Jill about the image. 

The full dataset is given by

\begin{equation}\label{eq:dataset}
    \mathcal{D} = \{(C_i,Y_i, e_i) : i \in \{1,2,...,180\}\}
\end{equation}

where $e_i$ is the metadata of the image $Y_i$ and contains valuable information useful for evaluation purposes. 

\subsection{Data Collection and Statistics}
We collected the images from three different sources: unique (not publicly available), Pinterest, and Flickr. We have tried to minimize the number of images from the two latter sources, because it is reasonable to assume that existing pretrained models have seen them during training. The images are pre-classified into six categories: cartoon, nature, painting, product, animal, and human, with each category containing 30 samples.

\begin{figure}[h]
    \centering
    \includegraphics[width=0.4 \textwidth, angle =270]{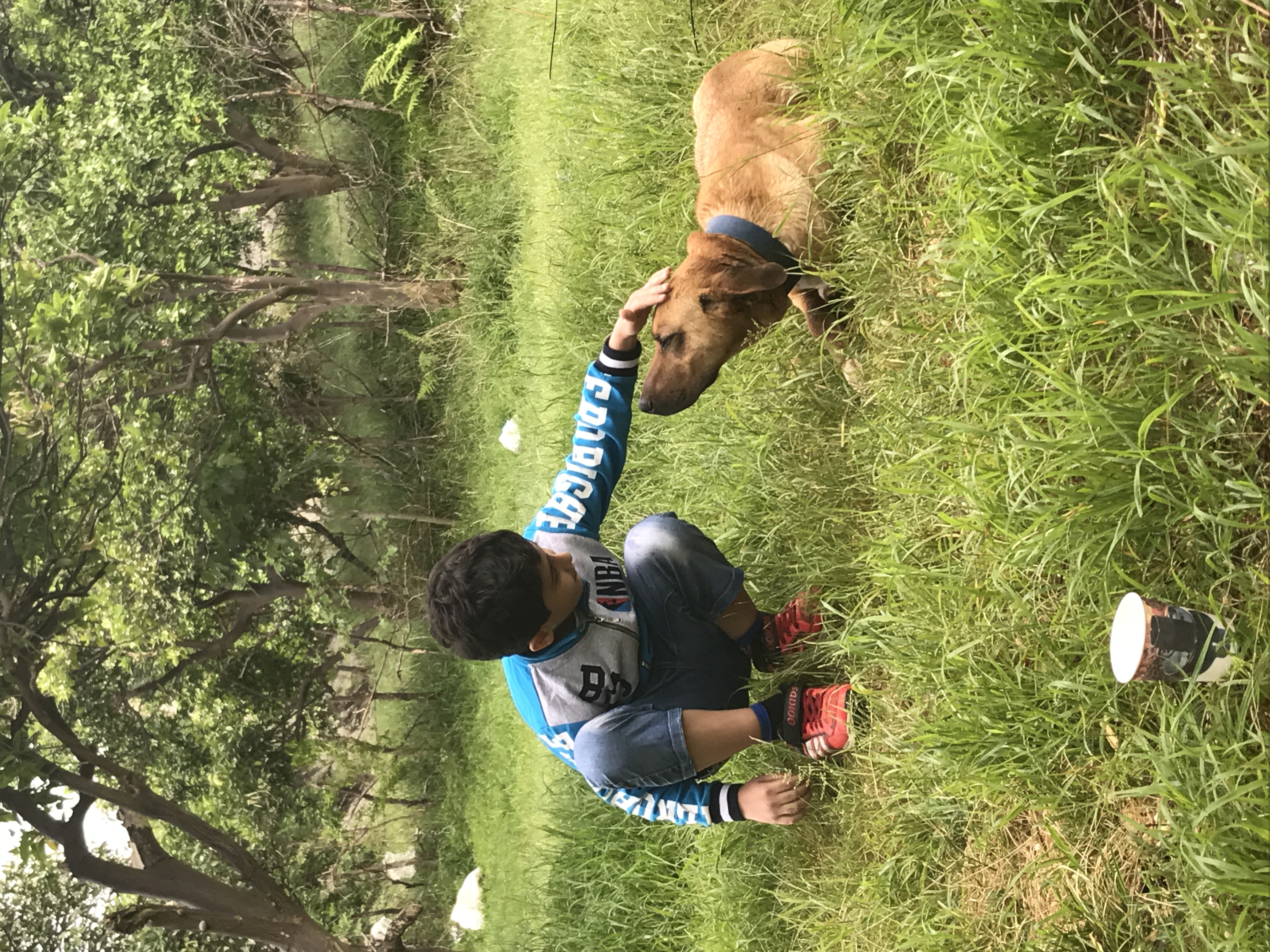}
    \caption{A sample image from the \textit{ConvGenVisMo} dataset.}
    \label{fig:img_sample}
\end{figure}

\begin{figure*}[t]
\centering
    \includegraphics[width=\columnwidth, height=1.4 \textwidth]{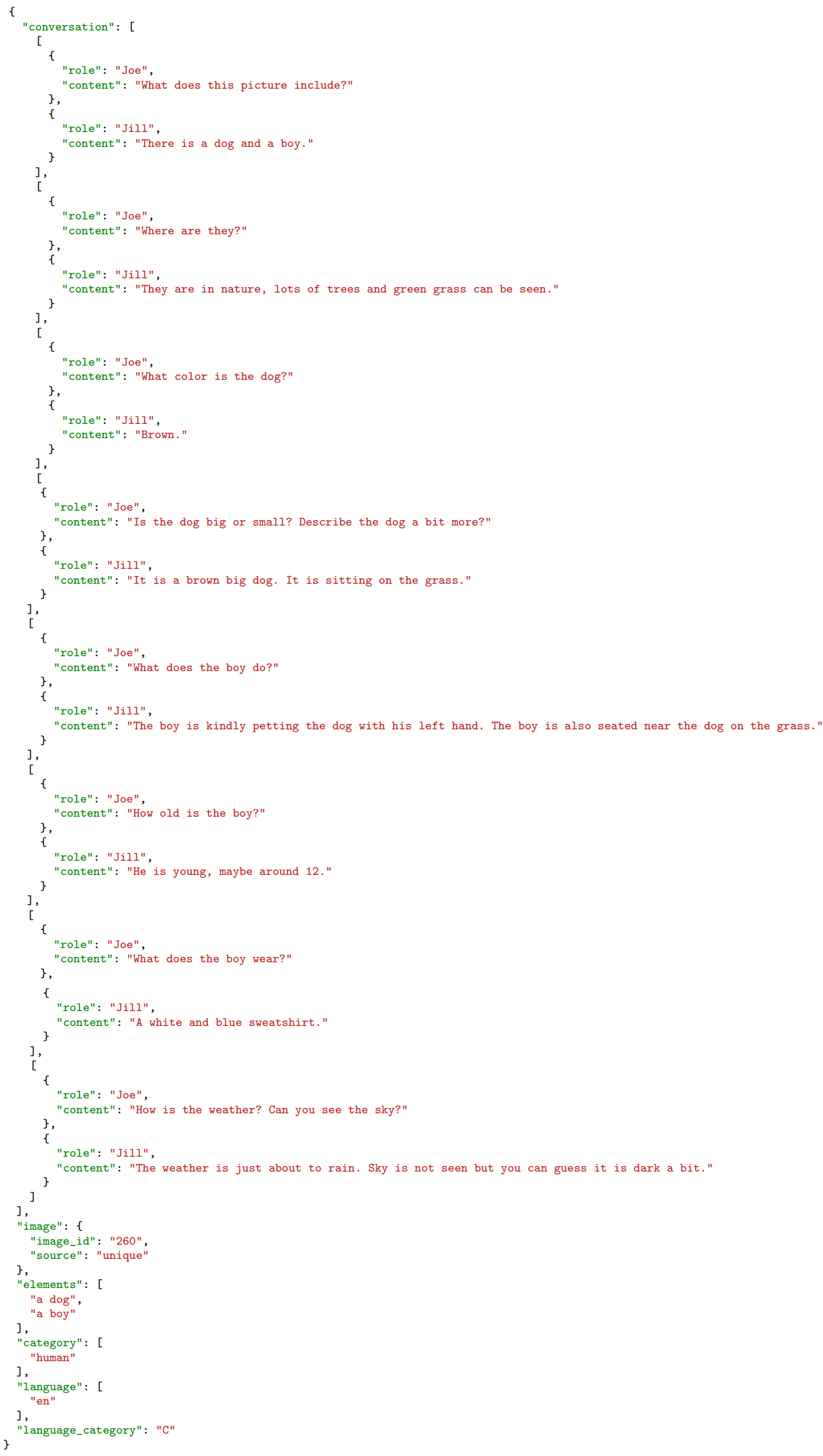}
    \caption{A sample conversation for the image in Figure~\ref{fig:img_sample}.}
    \label{metadata}
\end{figure*}

A sample image and its corresponding conversation are shown in Figure~\ref{fig:img_sample} and Figure~\ref{metadata} respectively.

\begin{figure*}[t]
    \centering
    \begin{subfigure}[t]{0.40\textwidth}
        \includegraphics[width=1.0 \textwidth, height=0.8 \textwidth]{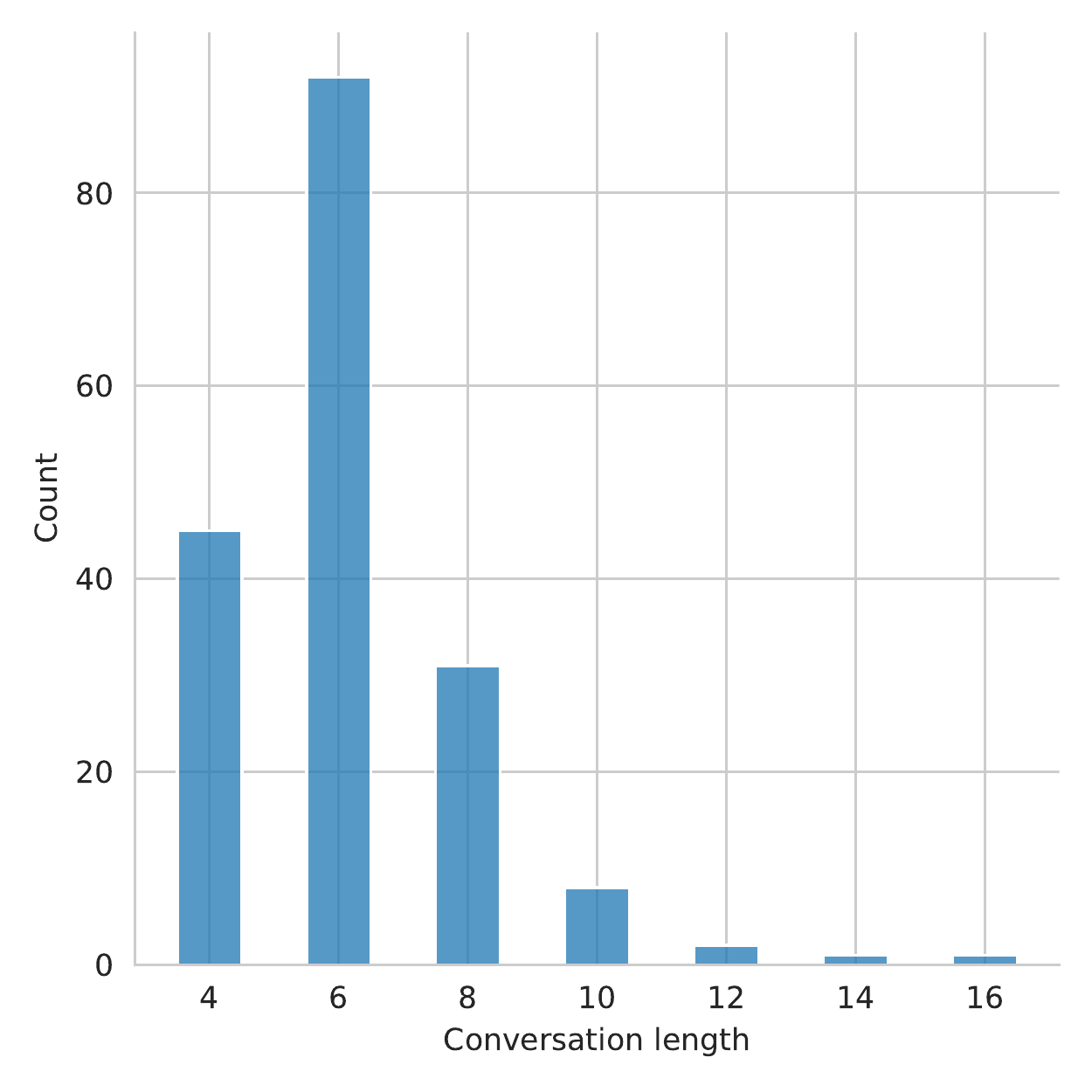}
        \caption{Conversation length distribution}
        \label{fig:conv_dist}
    \end{subfigure}%
    \begin{subfigure}[t]{0.40\textwidth}
        \includegraphics[width=1.0 \textwidth, height=0.8 \textwidth]{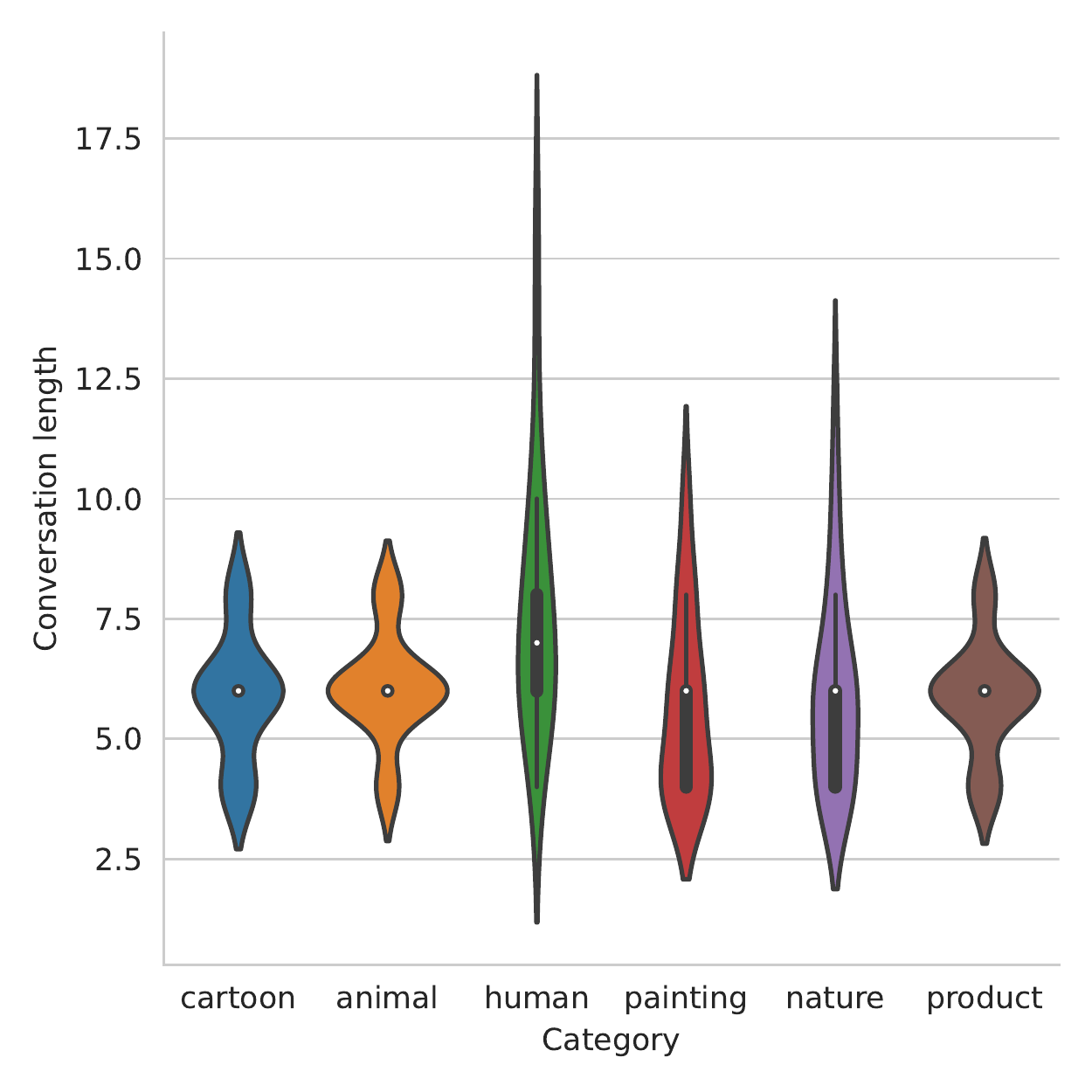}
        \caption{Categorical conversation length distribution}
        \label{fig:cat_conv_dist}
    \end{subfigure}
    \caption{Conversation-level statistics of the\textit{ConvGenVisMo} dataset.}
    \label{fig:conv_stats}
\end{figure*}

Figures~\ref{fig:conv_dist}-\ref{fig:cat_conv_dist} show the distribution statistics of conversations in the dataset. Images categorized as `Human' are the most common, and are widely spread across different conversation lengths. This is because most images in the dataset have been captured from mobile phones. In contrast, the `Product' category is the most uncommon, and has the smallest spread across conversation lengths.

\begin{figure*}[t]
    \centering
    \begin{subfigure}[t]{0.40\textwidth}
        \includegraphics[width=1.0 \textwidth]{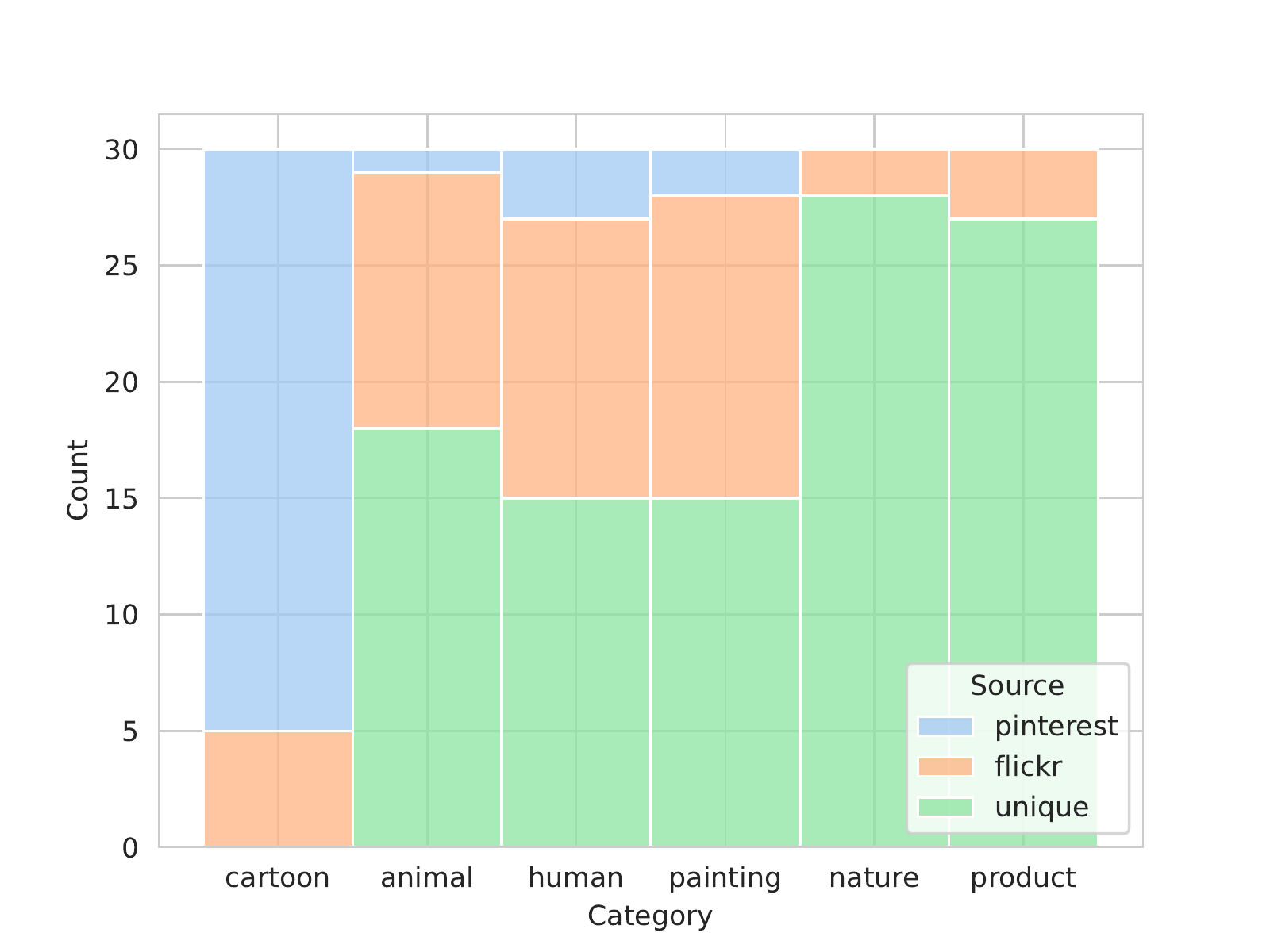}
        \caption{Categorical image source distribution}
        \label{fig:cat_img_src_dist}
    \end{subfigure}%
    \begin{subfigure}[t]{0.40\textwidth}
        \includegraphics[width=1.0 \textwidth]{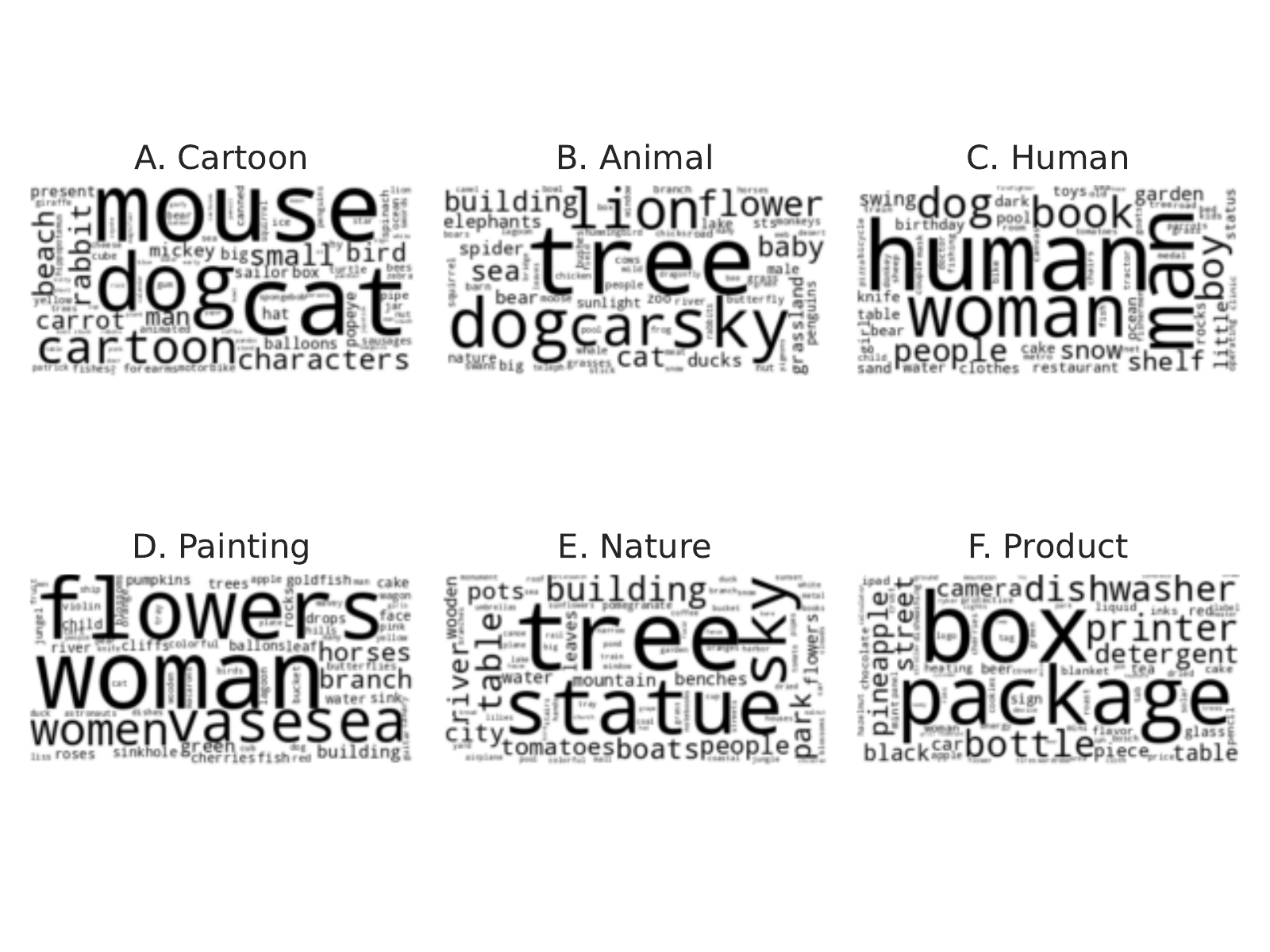}
        \caption{Categorical image element word cloud}
        \label{fig:cat_img_element_wc}        
    \end{subfigure}   
    \caption{Image-level statistics of the \textit{ConvGenVisMo dataset.}}
    \label{fig:image_stats}
\end{figure*}

Figures~\ref{fig:cat_img_src_dist}-\ref{fig:cat_img_element_wc} shows the distribution statistics of images in the dataset. The most common image source is `unique', which represents private, unseen images taken from personal mobile phones (and donated to us voluntarily). The wordcloud shown here is generated from visual element annotation of the images in each category.

\begin{figure*}[t]
    \centering
    \includegraphics[width=0.6 \textwidth]{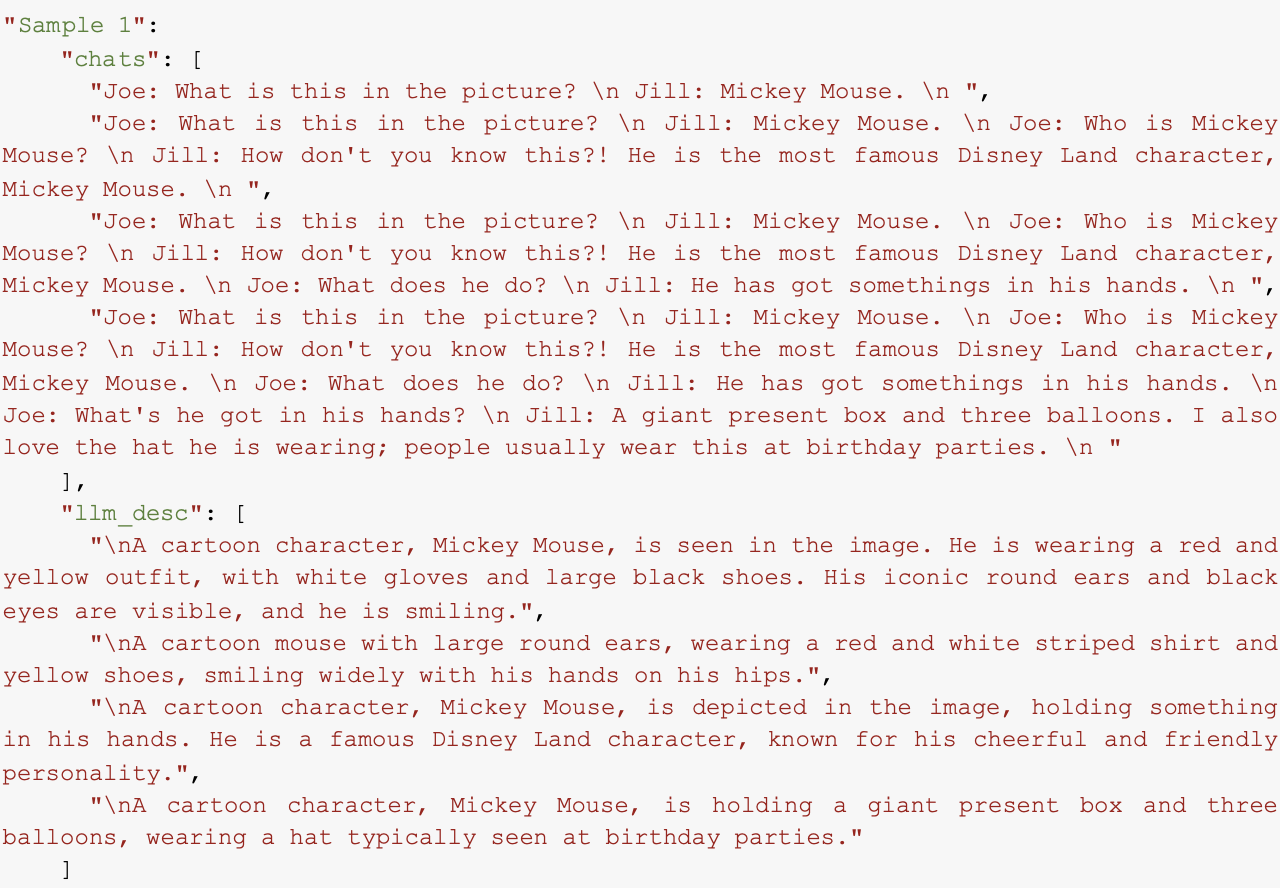}
    \caption{An example of ChatGPT input and output.}
    \label{chatgpt_input_output}
\end{figure*}

Figure~\ref{chatgpt_input_output} illustrates the input and output of ChatGPT for each hop of an example conversation. Each element of the \texttt{chats} array shows a hop. Each element of the \texttt{llm\_desc} array shows the generated summary of the conversation upto the corresponding hop.

\section{\textit{ConvGenVisMo} Evaluation Metrics}
%Many emerging researches are still being conducted on this topic. 

Automatic performance evaluation of CGVMs is an open research question, with no clear metrics proposed to date. 
In this paper, we explore several existing metrics from the literature on conversational models and visual question answering. We also propose a novel element presence score.

\subsection{Image Quality Assessment (IQA) Metrics}
Assessing the quality of generated images remains a crucial challenge in the field of generative vision models. To this end, numerous metrics have been proposed, including image quality assessment (IQA) metrics, which are categorized into `No Reference' (NR) or `Full Reference' (FR) metrics.

NR-IQA metrics are generally useful when reference images are not available, but we can still use them in our study to capture the intrinsic quality of the CGVM-generated images. We use \textbf{Blind/Referenceless Image Spatial Quality Evaluator (BRISQUE)} \cite{mittal2012no}, which is a popular NR metric.

FR-IQA metrics are used when reference images to are available for comparison, %In the \textit{ConvGenVisMo} dataset, in view of the fact that there are the reference images to contrast with the created images, FR-IQA is an excellent alternative. 
and can be categorized into two types: semantic and computational. Computational FR-IQA metrics use algorithmic approaches to compare the generated images with reference images. %To evaluate the performance of CGVMs, computational FR-IQA criteria provide the facilities to analyze the differences between the two reference and generated images and calculate a quantitative measure of their similarity or dissimilarity.
%Common computational FR-IQA metrics include: PSNR \cite{1284395}, UQI \cite{wang2002universal}, %\textit{NQM}, \textit{FID} \cite{heusel2017gans}, and SSIM \cite{1284395}. 
\textbf{Peak Signal-to-Noise Ratio (PSNR)}~\cite{1284395} is a well-known FR-IQA metric that calculates the difference between a signal's maximal strength and the power of corrupting noise that degrades the accuracy of its representation.
In our case, reference image can be treated as signals and the CGVM generated image as the noise. For a given reference image $Y$ and a generated image $\hat{Y}$, PSNR is given by:

\begin{equation}
  \text{PSNR}(Y,\hat{Y}) = \log_{10}{\left(\frac{R^2}{\text{MSE}(Y,\hat{Y})}\right)}
\end{equation}
where \textit{R} is the highest possible pixel value for an image (e.g., 255 for an 8-bit image) and the MSE is the Mean Squared Error given by:

\begin{equation}\label{eqn:psnr}
  \text{MSE}(Y,\hat{Y})= \frac{1}{MN}\sum_{i=1}^M \sum_{j=1}^N(Y[i,j]-\hat{Y}[i,j])
\end{equation}

Here, $M$ and $N$ refer to the number of rows and columns of an image, respectively. The indices $i$ and $j$ represent the row and column indices, respectively. A higher PSNR value indicates that the CGVM's output is more similar to the reference image, because their pixel values are closer to each other. 

Another common FR-IQA metric is \textbf{Universal Quality Index (UQI)}~\cite{wang2002universal}. It compares the correlation (spatial arrangement of pixels), luminance (average pixel intensity), and contrast values of a generated image with those of the reference image. 
UQI is given by

\begin{equation}\label{eq_UQI}
    \text{UQI}(Y,\hat{Y})=\frac{\sigma_{Y \hat{Y}}}{\sigma_Y \sigma_{\hat{Y}}} \cdot \frac{2 \bar{Y} \bar{\hat{Y}}}{(\bar{Y})^2+(\bar{\hat{Y}})^2} \cdot \frac{2 \sigma_Y \sigma_{\hat{Y}}}{\sigma_Y^2+\sigma_{\hat{Y}}^2}
\end{equation}

where $\bar{Y}$ and $\bar{\hat{Y}}$ are defined as $\bar{Y}=\frac{1}{N} \sum_{i=1}^N Y_i$ and $\bar{\hat{Y}}=\frac{1}{N} \sum_{i=1}^N \hat{Y}_i$ respectively, $\sigma_{Y}^{2}$ and $\sigma_{\hat{Y}}^{2}$ denote the variances of $Y$ and $\hat{Y}$ respectively, and $\sigma_{Y\hat{Y}}$ is the covariance of $Y$ and $\hat{Y}$.

\textbf{Structural Similarity Index (SSIM)}~\cite{1284395} is another popular FR-IQA metric, which is predicated on the notion that changes in structural information are more perceptible to the human visual system than changes in pixel values alone. SSIM allows us to take into account the structural similarity of two images. It computes luminance (\textit{l}), contrast (\textit{c}) and structure (\textit{s}) similarities as follows:

\iffalse 
\begin{equation}
SSIM(Y,\hat{Y})=\frac{\left(2 \mu_Y \mu_{\hat{Y}}+c_1\right)\left(2 \sigma_{Y {\hat{Y}}}+c_2\right)}{\left(\mu_Y^2+\mu_{\hat{Y}}^2+c_1\right)\left(\sigma_Y^2+\sigma_{\hat{Y}}^2+c_2\right)}
\end{equation}
\fi

\begin{subequations}\label{eq_lcs}
\begin{equation}\label{eq_l}
    l(Y,\hat{Y})=\frac{2 \mu_Y \mu_{\hat{Y}}+c_1}{\mu_Y^2+\mu_{\hat{Y}}^2+c_1}
\end{equation}
\begin{equation}\label{eq_c}
    c(Y,\hat{Y})=\frac{2 \sigma_Y \sigma_{\hat{Y}}+c_2}{\sigma_Y^2+\sigma_{\hat{Y}}^2+c_2}
\end{equation}
\begin{equation}\label{eq_s}
    s(Y,\hat{Y})=\frac{\sigma_{Y {\hat{Y}}}+c_3}{\sigma_Y \sigma_{\hat{Y}}+c_3}
\end{equation}
\end{subequations}

where $\mu_{Y}$ and $\mu_{\hat{Y}}$ are the pixel sample means of $Y$ and $\hat{Y}$, respectively. To stabilize the division with a weak denominator, three variables $c_{1}=(k_{1}L)^{2}$,  $c_{2}=(k_{2}L)^{2}$, and $c_{3}=c_{2}/2$ are used. $k_1$, $k_2$ are hyper-parameters and $L$ is the dynamic range for pixel values. Based on the Eq.~\ref{eq_lcs},  SSIM is defined as

\begin{equation}\label{eq_SSIM}
\text{SSIM}(Y,\hat{Y})=l(Y,\hat{Y})^\alpha \cdot c(Y,\hat{Y})^\beta \cdot s(Y,\hat{Y})^\gamma
\end{equation}
where $\alpha$, $\beta$ and $\gamma$ are the weights to combine the three components.

 Note that both SSIM and UQI use luminance and contrast in their evaluation of image quality, however they do so in different ways and for different purposes. In SSIM, the contrast similarity signifies the standard deviation of the pixel values, and the luminance comparison quantifies the average brightness of the images. The similarity of the edges and textures of the reference and generated images is measured by the structural comparison. 

It is important to note that each aforementioned metric has advantages and limitations, thus we use multiple such metrics in our evaluation framework. 

Computational IQA metrics have a serious limitation: they are often divorced from human perception. %, thereby they lose touch with the individualized experiences and viewpoints that are fundamental to human comprehension.  
For example, rotated images are identical for a human viewer, but computational IQA techniques will give them low similarity scores. Therefore, it is essential to develop alternative metrics which factor in human perception and judgement.
 
Semantic FR-IQA metrics provide a solution to the aforementioned issue by taking into account the human perception of image quality. Semantic FR-IQA metrics incorporate features such as sharpness, color accuracy, and contrast to evaluate the image quality. \textbf{CLIP similarity score} \cite{DBLP:journals/corr/abs-2103-00020} is an important metric in this category, which can be used to measure the semantic alignment between two images in a high dimensional vector space. A higher CLIP score indicates higher similarity between two images. %It can be also used for semantic similarity measurement of two images. In our case, we use it as image semantic similarity metric. However, other advanced methods can also replace it but the overall goal is to use a semantic model to represent both images, generated and ground truth in a semantic space and evaluate their vector similarity. 
CLIP score is given by 
\begin{equation} \label{eq:clip_score}
    \text{CLIP}_{score}(Y, \hat{Y}) = cos(\text{CLIP}(Y), \text{CLIP}(\hat{Y}))
\end{equation}
where CLIP() is a transformation function that maps an image to a semantic vector space.

While CLIP score offers advantages in evaluating image quality by capturing semantic information, it may have limitations in accurately considering the presence of specific objects in the images. Therefore, there is a need for evaluation metrics that specifically measure the presence of elements or objects in images.

\subsection{Element Presence Scores}
For two images to be considered similar, they must have common elements (e.g. people or objects). We capture this notion with Element Presence Scores.

\textbf{Element Presence Precision (EPPr)} is defined as number of overlapping elements present in the generated image and ground-truth, divided by number of elements in the generated image. \textbf{Element Presence Recall (EPRe)}, on the other hand, is defined as number of overlapping elements present in the generated image and ground-truth, divided by number of elements in the ground-truth image. Intuitively, these scores combine the notion of element presence with the regular definitions of precision and recall. 
\begin{equation}\label{eq:epp}
    \text{EPPr}(Y, \hat{Y}) = \frac{\text{Elements}(\hat{Y}) \cap \text{Elements}(Y)}{\text{Elements}(\hat{Y})}
\end{equation}
\begin{equation}\label{eq:erp}
    \text{EPRe}(Y, \hat{Y}) = \frac{\text{Elements}(\hat{Y}) \cap \text{Elements}(Y)}{\text{Elements}(Y)}
\end{equation}

\begin{figure}[t]
    \centering
    \includegraphics[width=\columnwidth]{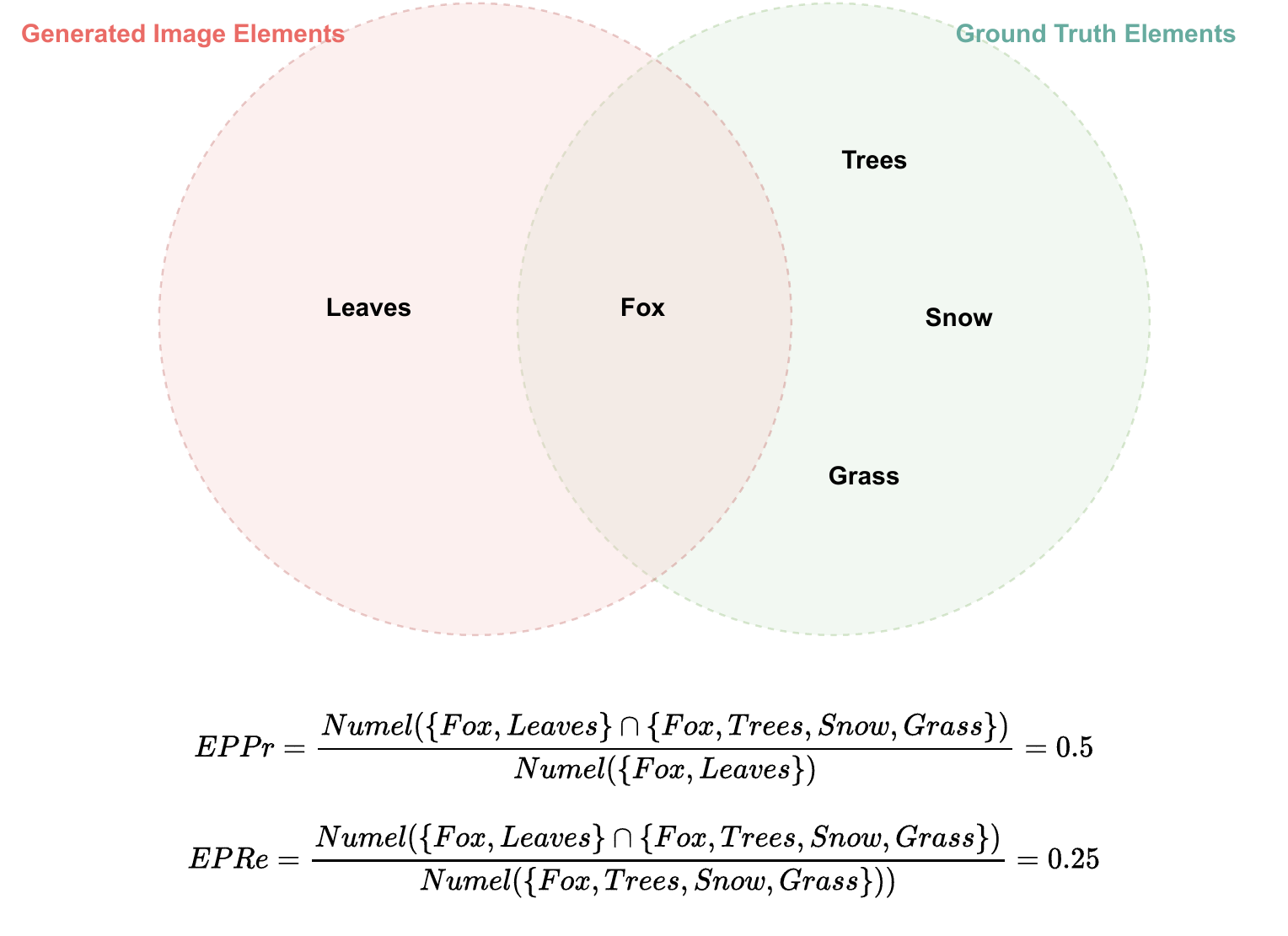}
    \caption{EPPr and EPRe for hop-2 for example in Fig. \ref{fig:example}; Also known as: EPPr@2 and EPRe@2.}
    \label{fig:PrRe}
\end{figure}

We can similarly define \textbf{Element Presence F1 (EPF1)} score. For element detection in images, we can use out-of-the-box automated object detection methods, or have humans annotate harder examples.
Figure~\ref{fig:PrRe} shows the EP computations on an example image.

%As shown, Element Presence Scores are computed for each sample separately, and can be combined via macro or micro average. The micro-average can be calculated for different elements separately, and thus more finegrained insights can be derived about the generative model's behaviour. For example, with micro-average of EPF1 we can understand how robust a specific model is in generating images of humans. 

\begin{figure}[t]
    \centering
    \includegraphics[width=\columnwidth]{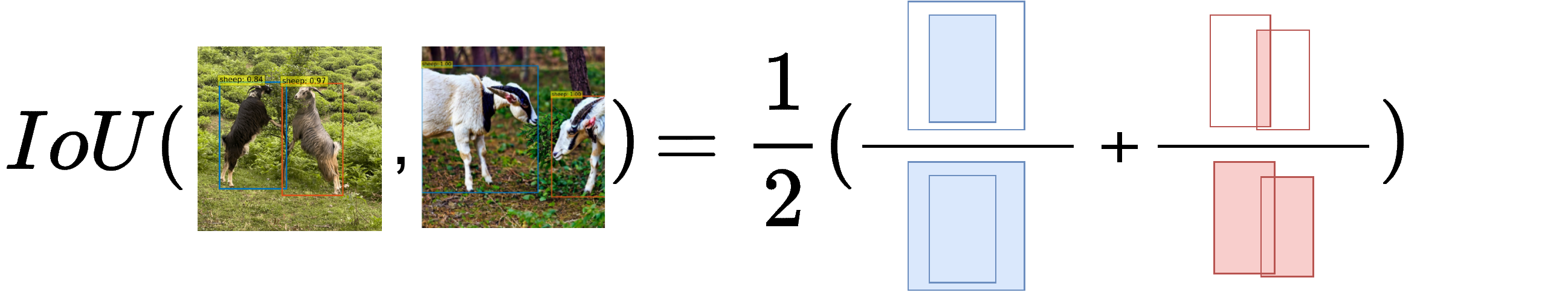}
    \caption{IoU metric for ground truth image (left) and generated image (right).}
    \label{fig:iou}
\end{figure}

\textbf{Intersection over Union (IoU)} is another metric that can be used to calculate the presence of objects and their respective position using an object detection method. %For this case, we will assume that not only a single image, but two images are present, ground-truth image and generated one. This metric applies object detection on both of these images and according to objects that are present in the ground-truth image, IoU and element presence score can be calculated. 
It is given by

\begin{equation}\label{eq:iou}
    \mathrm{IoU}(Y,\hat{Y}) = \frac{1}{N} \sum_{e=1}^{N} \frac{B_{e}(Y) \cap B_{e}(\hat{Y})}{B_{e}(Y) \cup B_{e}(\hat{Y})}
\end{equation}

where $N$ is the number of objects in the ground truth image and $e$ represents each one of these objects. $B$ is a function that gives the bounding box of each object $e$ in the image. Figure~\ref{fig:iou} shows the intuition of IoU computation on example images.

IoU is mainly useful in cases where the position of objects in the generated image is important, however it is possible that objects don't not appear in either of the images. Given this fact, we propose three variants of: \textbf{Common-IoU}, \textbf{Precision-IoU}, and \textbf{Recall-IoU}. Common-IoU is calculated in the same way but only for objects that are present in both of the images. Precision-IoU  calculates IoU for objects present in the generated image and when an object only appears in the generated but not in the original image, the score for that specific object is zero. Similarly, Recall-IoU takes only the original image objects into account and gives a zero score for that specific object if it is not present in the generated image. %The aggregation of object-based scores for all these three is an average function. Meaning that, 
For each object, we compute IoU for all of the instances of that object in the image and then average the result over all objects.

\subsection{Top Conversation Hops}

We further introduce the concept of @K for each evaluation metric described previously, which denotes the value of that metric upto the K'th conversation hop. An example is EPPr@K, read as Element Presence Precision at K.

In the same vein, we can define the notion of image generated at K, given by
%Breaking down equations \ref{eq:conversation} and \ref{eq:dataset}, we can see each conversation hop as illustrated in Eq. \ref{eq:hop}. Function $f$ is the conversational generative vision model. This model can take the conversation up to a specific hop and generate an image such as $\hat{Y}_K$.
\begin{equation}\label{eq:hop}
    \hat{Y}_K = f(\{(M_{Joe}, M_{Jill})_t : 1 \leq t \leq K \})
\end{equation}

\section{Experimental evaluation}
Here we provide details of the CGVM evaluation process. %In the first subsection, we describe the evaluation settings and the challenges we encountered during the evaluation process. The second subsection presents the results for the different groups of automated evaluation metrics.

\subsection{Evaluation process and settings}
%This section aims to provide the reader with more detailed information about the specific models used for each part of the evaluation, along with the corresponding prompts and any additional relevant details. Furthermore, this section offers a comprehensive understanding of the challenges encountered during the evaluation process. 

We demonstrate the \textit{ConvGenVisMo} evaluation framework on a CGVM which is a combination of ChatGPT (``text-davinci-003'' conversation model) and DreamStudio (``stable-diffusion-xl-beta-v2-2-2'' vision model). %Ofcourse, these can be substituted with any other conversational and/or vision models.

The overall evaluation process consists of three steps: 

\begin{enumerate}
\item \textbf{Generate conversation summaries:} For each conversation sample in the \textit{ConvGenVisMo} dataset, we generate a textual summary of the conversation upto each hop, using ChatGPT. For example, if a conversation sample has 2 hops, we generate two summaries for it. The first summary is for the first hop, and the second summary is of both hops. Figure~\ref{chatgpt_input_output}  illustrates the input and output of ChatGPT for each hop of an example conversation. 

We use the following prompt for ChatGPT:

``\texttt{Below is a conversation between Joe and Jill, about an image. Use this conversation to generate a description of the image, such that it can be given as input to a text-to-image model as a prompt.}''

%While carrying out this step, we observed that DreamStudio does not generate images for conversation samples that contain references to children. DreamStudio considers such content as sensitive material and prohibits its generation. 

\item \textbf{Generate images for summaries:} For each summary of each conversation obtained in step 1, we generate an image using DreamStudio.

Figure \ref{fig:example} shows an example conversation and the respective image generated at each conversation hop.

\item \textbf{Comparison of generated images with ground truth:} For each conversation, we compare its associated generated images with the ground truth image using the automated evaluation metrics. We then aggregate the statistics of this analysis.

To ensure accurate computation of metrics, we perform a number of preprocessing steps on all the images. We standardize the size and format of each image, perform object detection on them using the DETR-100 algorithm  \cite{carion2020end}, and use linear interpolation to normalize the number conversation hops, so that they fall within the range $0$ to $1$. A hop value of $0$ represents the first conversation hop, while a value of $1$ corresponds to the last conversation hop. This allows meaningful comparisons between conversations that have varying numbers of hops.

\end{enumerate}

\subsection{Evaluation Results}
Here we present the results for the two groups of CGVM evaluation metrics. %As mentioned previously, these metrics were categorized into distinct groups based on the specific aspects being evaluated.

\subsection{CGVM Performance on IQA Metrics}
We organize our IQA results by the types of IQA metrics: NR-IQA, computational FR-IQA, and semantic FR-IQA. %This division allows for a comprehensive analysis of different aspects related to generated image quality.

\begin{figure}[t]
    \centering
    \includegraphics[width=0.35 \textwidth]{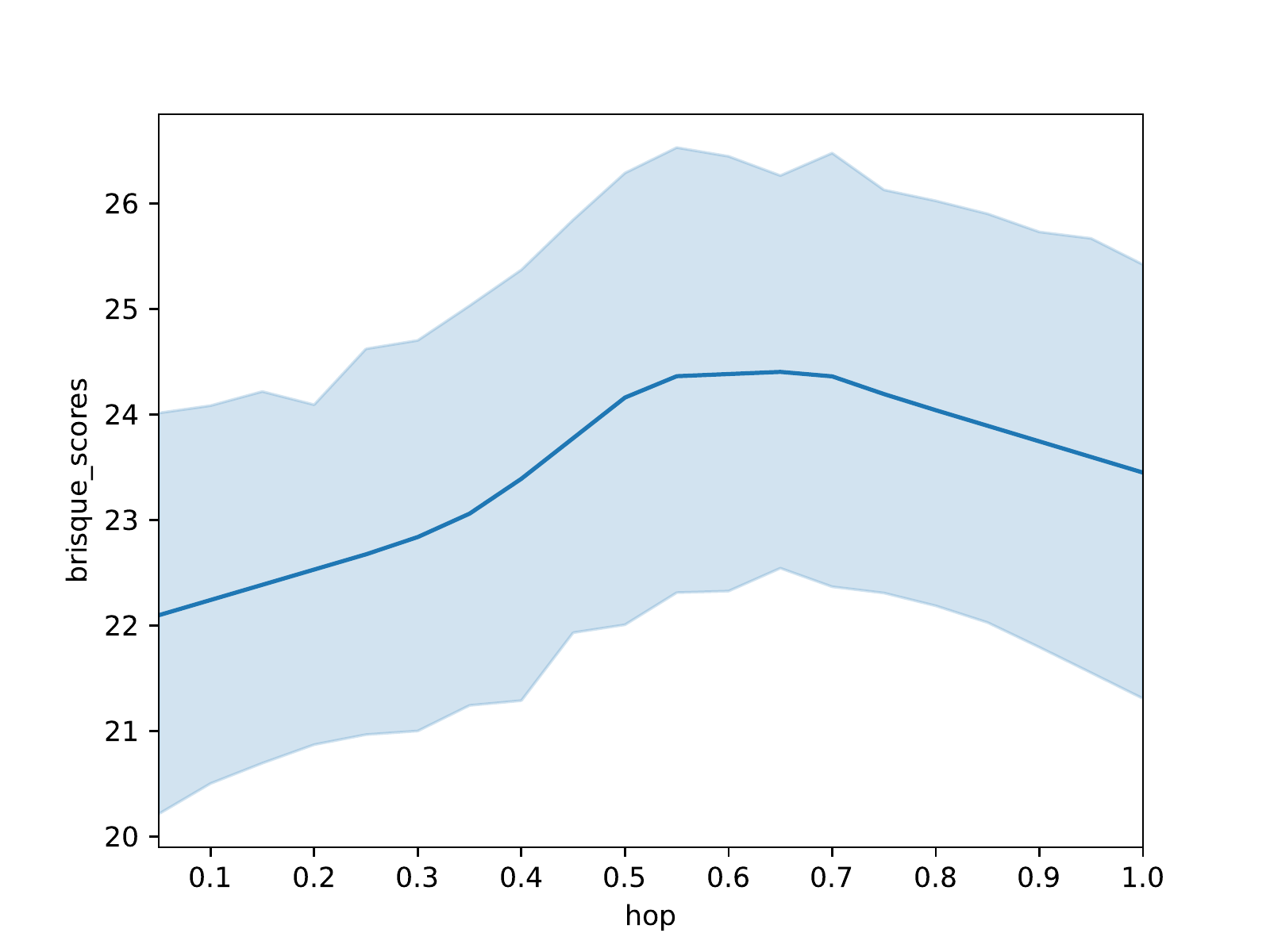}
    \caption{Model performance  based on NR-IQA metric: BRISQUE score.}
    \label{fig:brisque_score_re}
\end{figure}

\textbf{NR-IQA metric - BRISQUE:} %To evaluate the performance of CGVMs independently of reference images, we approach the task as an unsupervised problem. For this purpose, we utilize the BRISQUE score. 
    Figure~\ref{fig:brisque_score_re} illustrates the BRISQUE score corresponding to the number of hops. We see a somewhat linear trend, with a slight decrease after reaching a peak.  
    A lower BRISQUE score corresponds to better perceptual quality, so the fact that the scores remain within a narrow range (20--26) suggests that the created images have consistently high quality overall. This in turn suggests that the model is good at preserving the quality of the generated images.

\begin{figure*}[t]
    \centering
    \begin{subfigure}[t]{0.22\textwidth}
        \includegraphics[width=1.0 \textwidth]{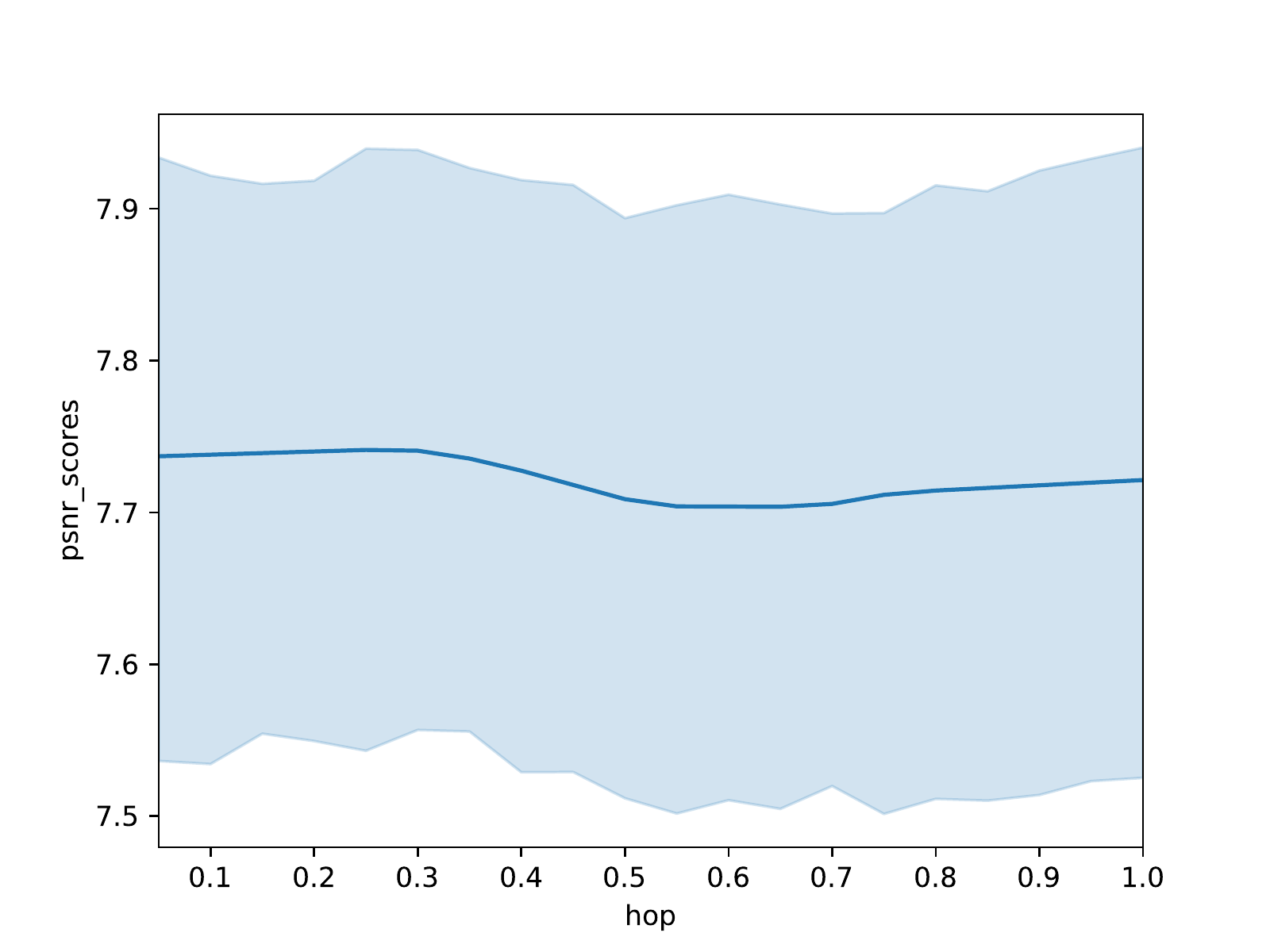}
        \caption{PSNR score(db)}
        \label{fig:psnr_score_re}        
    \end{subfigure}% 
    \begin{subfigure}[t]{0.22\textwidth}
        \includegraphics[width=1.0 \textwidth]{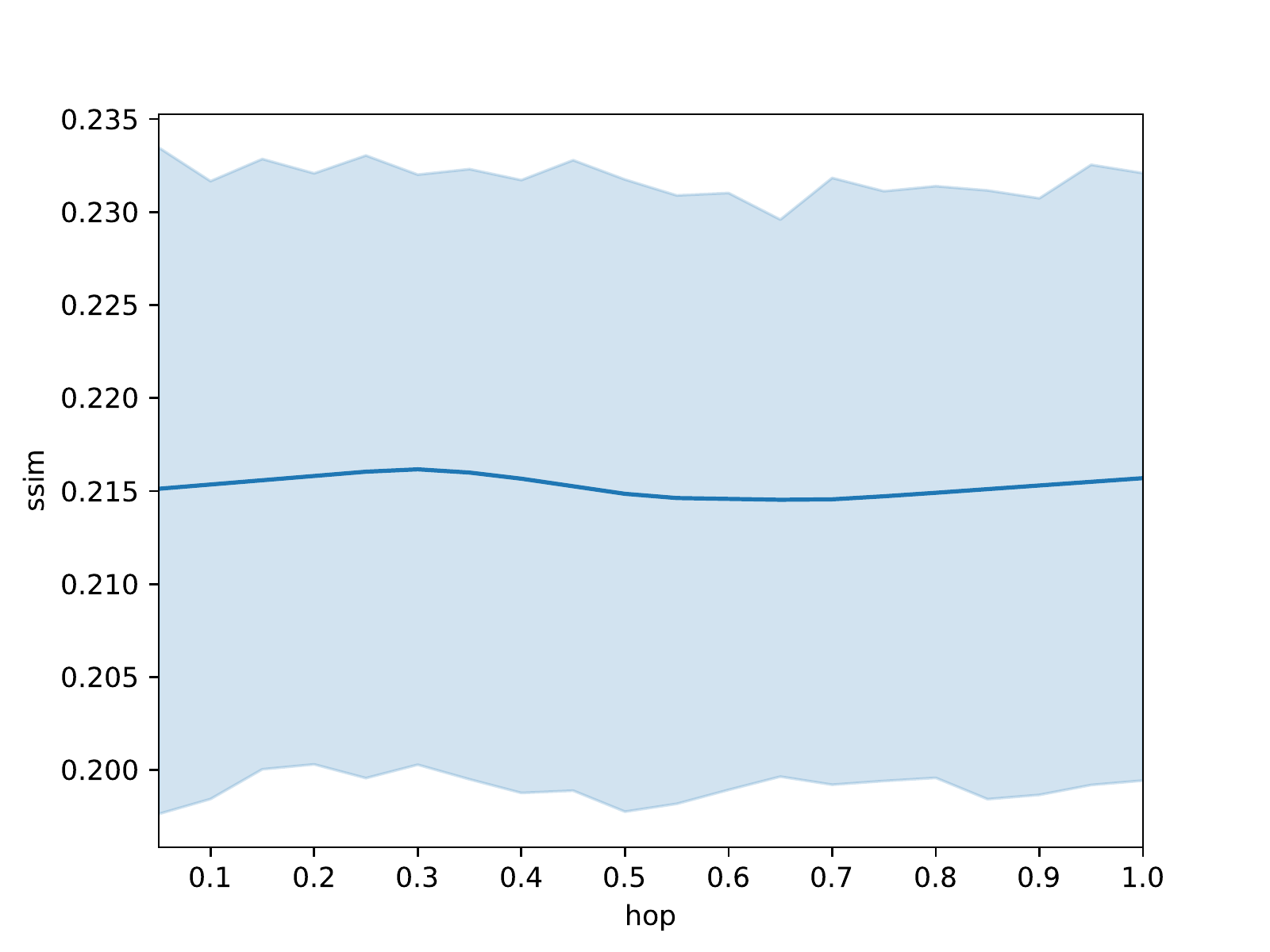}
        \caption{SSIM score}
        \label{fig:ssim_score_re}        
    \end{subfigure}%
    \begin{subfigure}[t]{0.22\textwidth}
        \includegraphics[width=1.0 \textwidth]{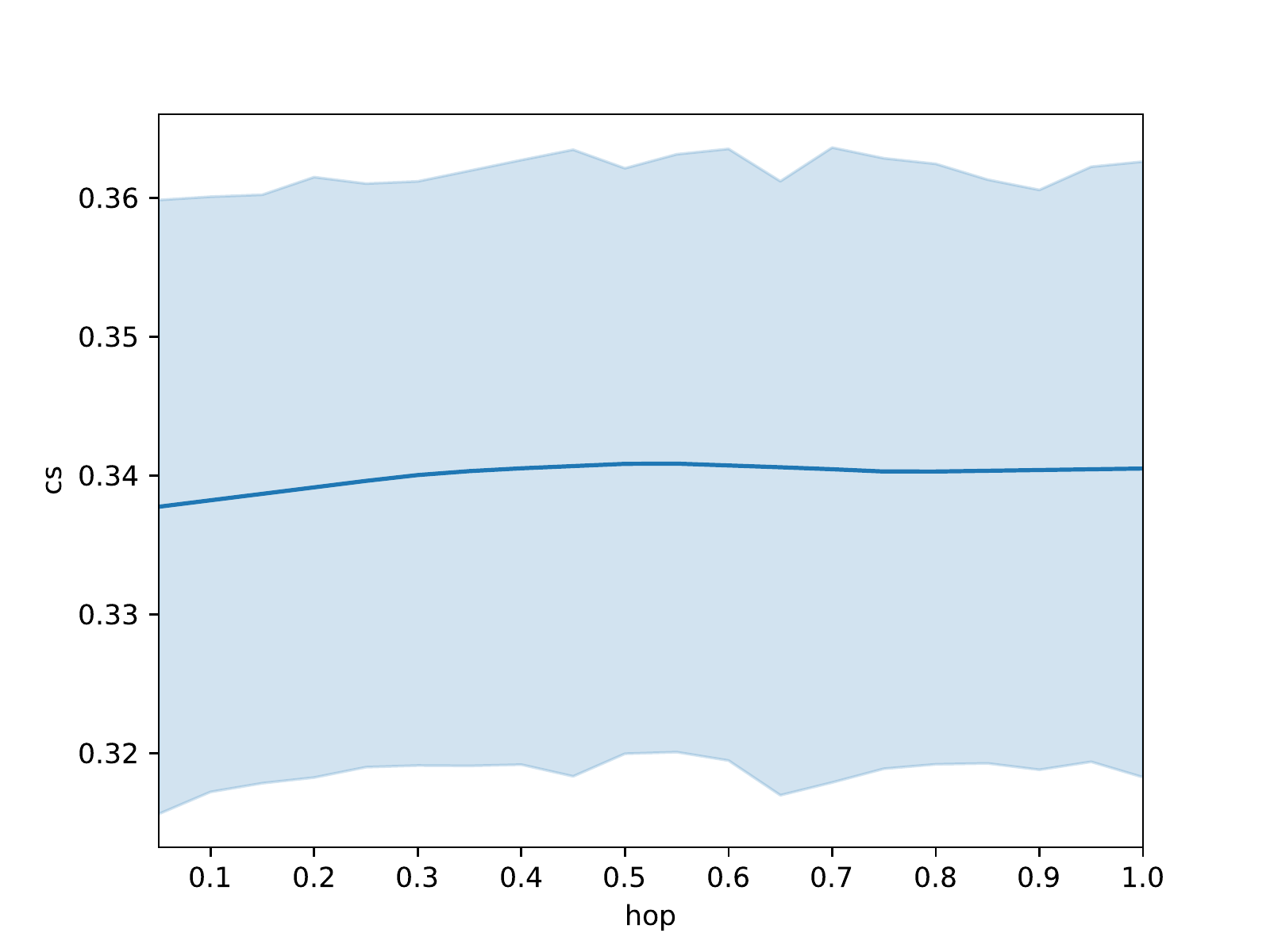}
        \caption{Contrast similarity component of SSIM}
        \label{fig:cs_score_re}        
    \end{subfigure}%
    \begin{subfigure}[t]{0.22\textwidth}
        \includegraphics[width=1.0 \textwidth]{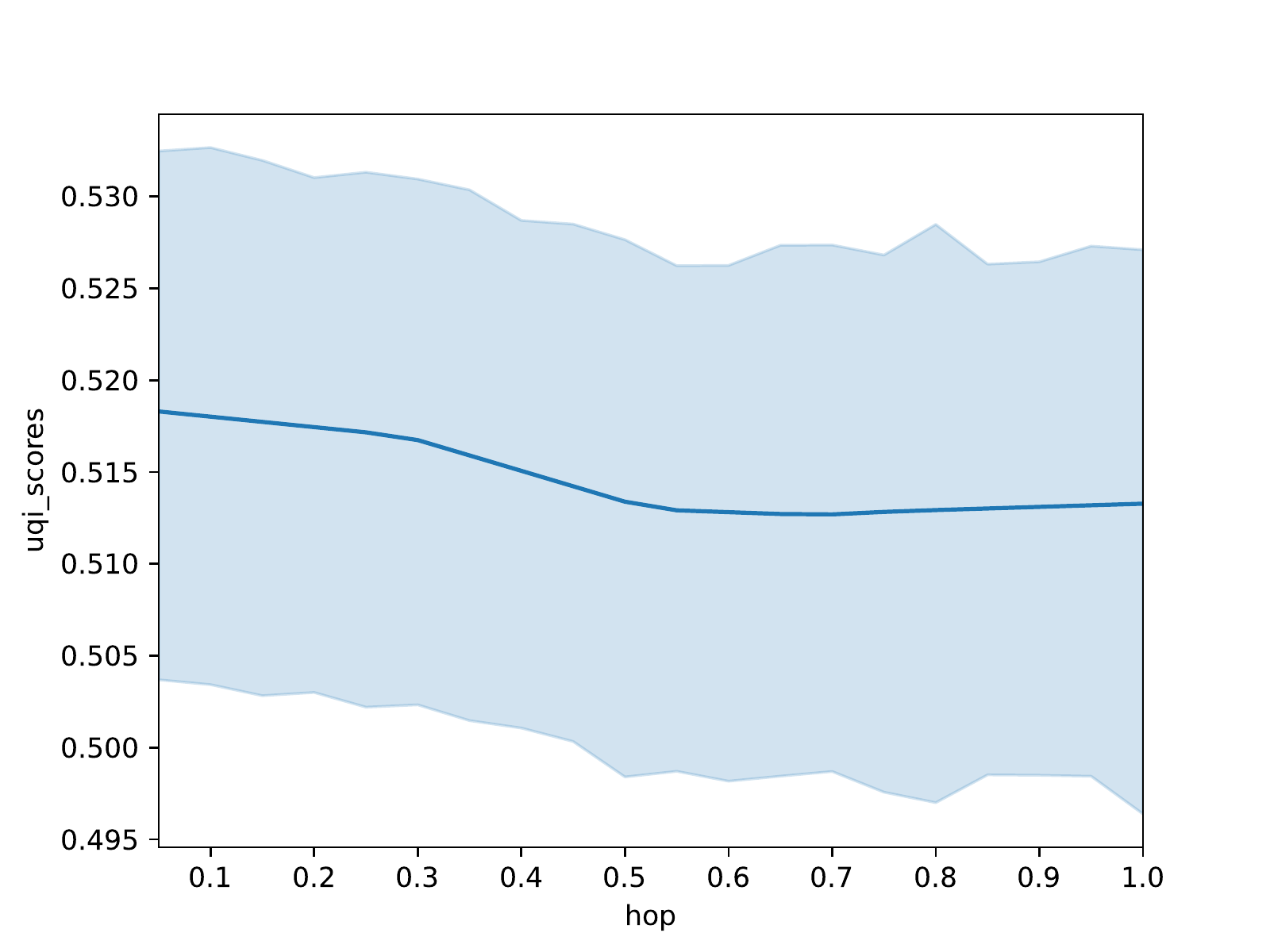}
        \caption{UQI score}
        \label{fig:uqi_score_re}        
    \end{subfigure}  
    \caption{Model performance based on computational FR-IQA metrics.}
    \label{fig:c_FR_IQA_score_re}
\end{figure*}

\textbf{Computational FR-IQA metrics - PSNR, SSIM, UQI:}   Figures~\ref{fig:psnr_score_re}-\ref{fig:uqi_score_re} show the PSNR, SSIM, and UQI scores. We observe that these scores yield low values with minimal variation. As new details are introduced in each hop,  they do not accurately capture the perceptual impact of the added details, especially if the changes are significant.  This observation aligns with the fact that PSNR, SSIM, and UQI primarily focus on measuring pixel-wise differences, structural similarities, and universal quality aspects, respectively, rather than being concerned with higher-level image semantic characteristics or the presence of specific objects.

\begin{figure*}[t]
    \centering
    \begin{subfigure}[t]{0.45\textwidth}
         \includegraphics[width=1.0 \textwidth]{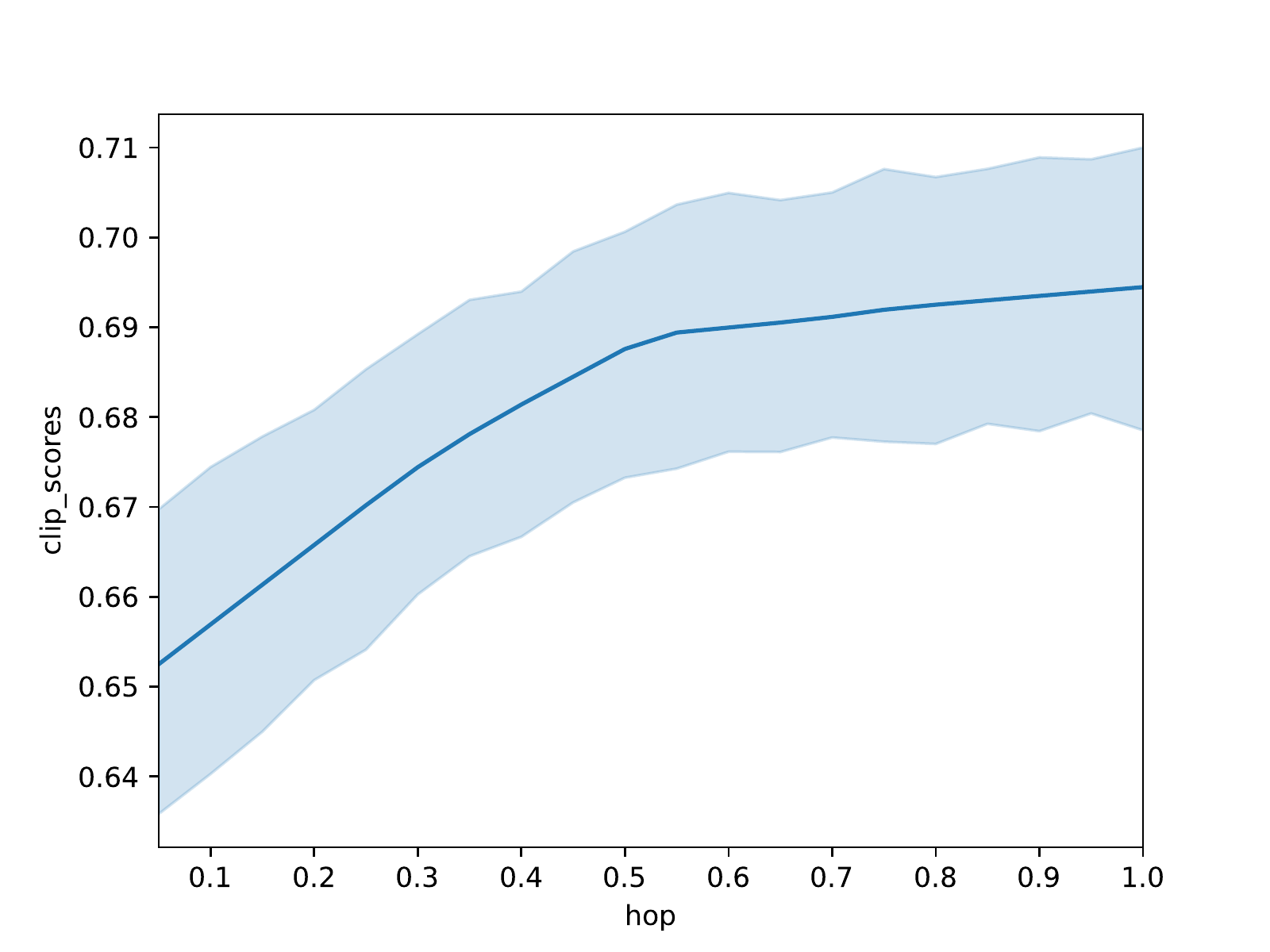}
        \caption{CLIP score based on hops}
        \label{fig:clip_score_re}      
    \end{subfigure}   
    \begin{subfigure}[t]{0.45\textwidth}
        \includegraphics[width=1.0 \textwidth]{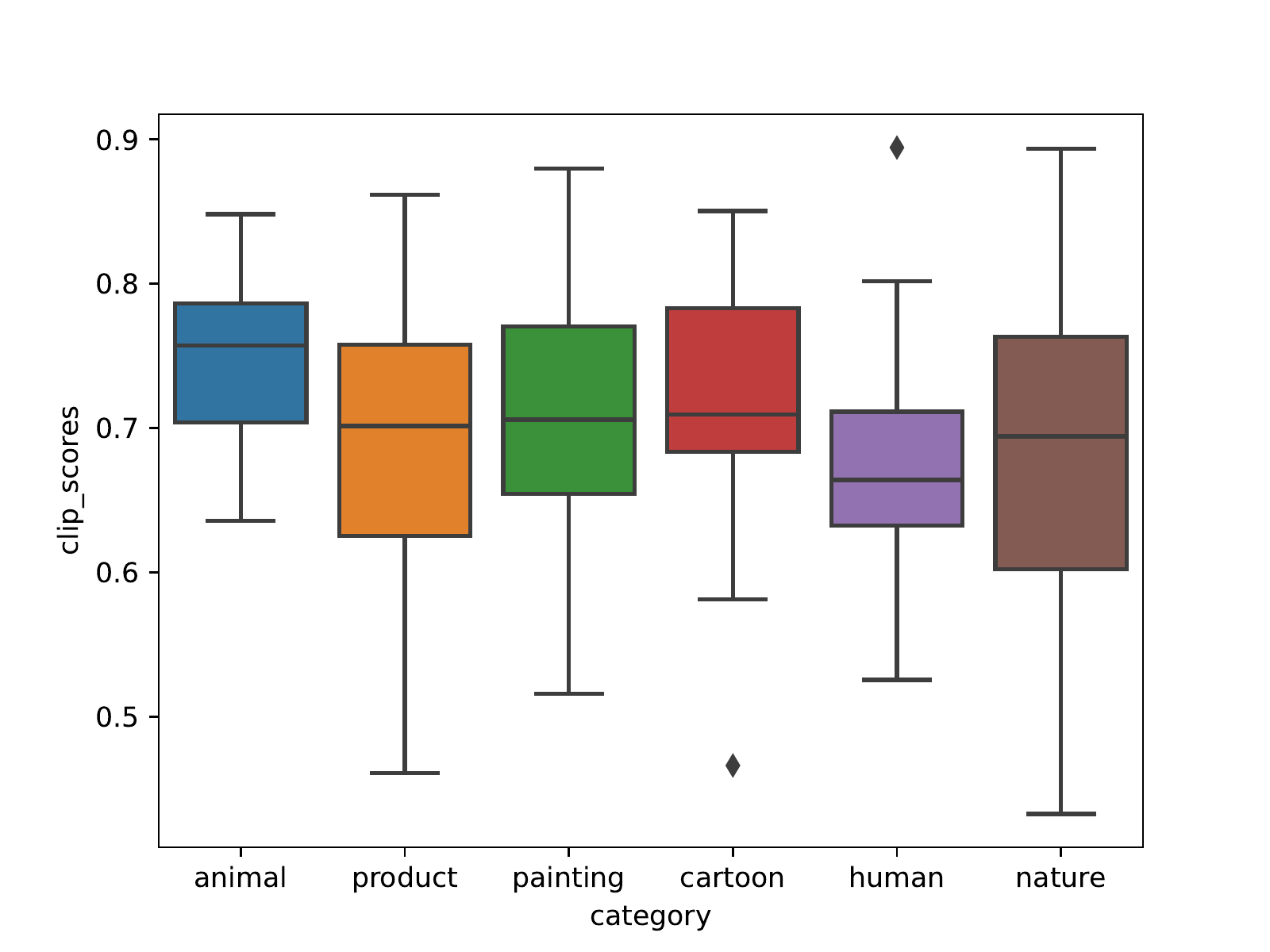}
        \caption{CLIP score based on each category for final step}
        \label{fig:clip_score_re_cat}
    \end{subfigure} 
    \caption{Model performance based on semantic FR-IQA metric: CLIP score.}
    \label{fig:results_clip_score}
\end{figure*}

\textbf{Semantic FR-IQA metric - CLIP Score:} Figure~\ref{fig:clip_score_re} demonstrates that as the number of hops increases, the CLIP score values also increase. This observation indicates that with each additional hop, more details are incorporated into the image, resulting in a higher similarity to the original image. In other words, the CLIP score accurately reflects the changes and improvements in the image as each step progresses. 

Figure~\ref{fig:clip_score_re_cat} provides a visual representation of the CLIP scores for different image categories. The animal category obtains the highest CLIP score, indicating a strong correlation between the generated images and the original images in this category. On the other hand, the human category exhibits the lowest CLIP score, suggesting a relatively weaker resemblance between the generated images and the original images in this category. The variations in CLIP scores across different categories highlight the varying performance of the CGVM in generating images related to different object categories.

\subsection{CGVM Performance on Element Presence Scores}
Recall that the Element Presence Scores provide insights into the model's ability to accurately generate objects and their spatial alignment with the ground truth objects in the images. On the other hand, the IoU score assesses the model's ability to accurately place the generated objects within the images.

\begin{figure*}
    \centering
    \begin{subfigure}[t]{0.33\textwidth}
        \includegraphics[width=1.0 \textwidth]{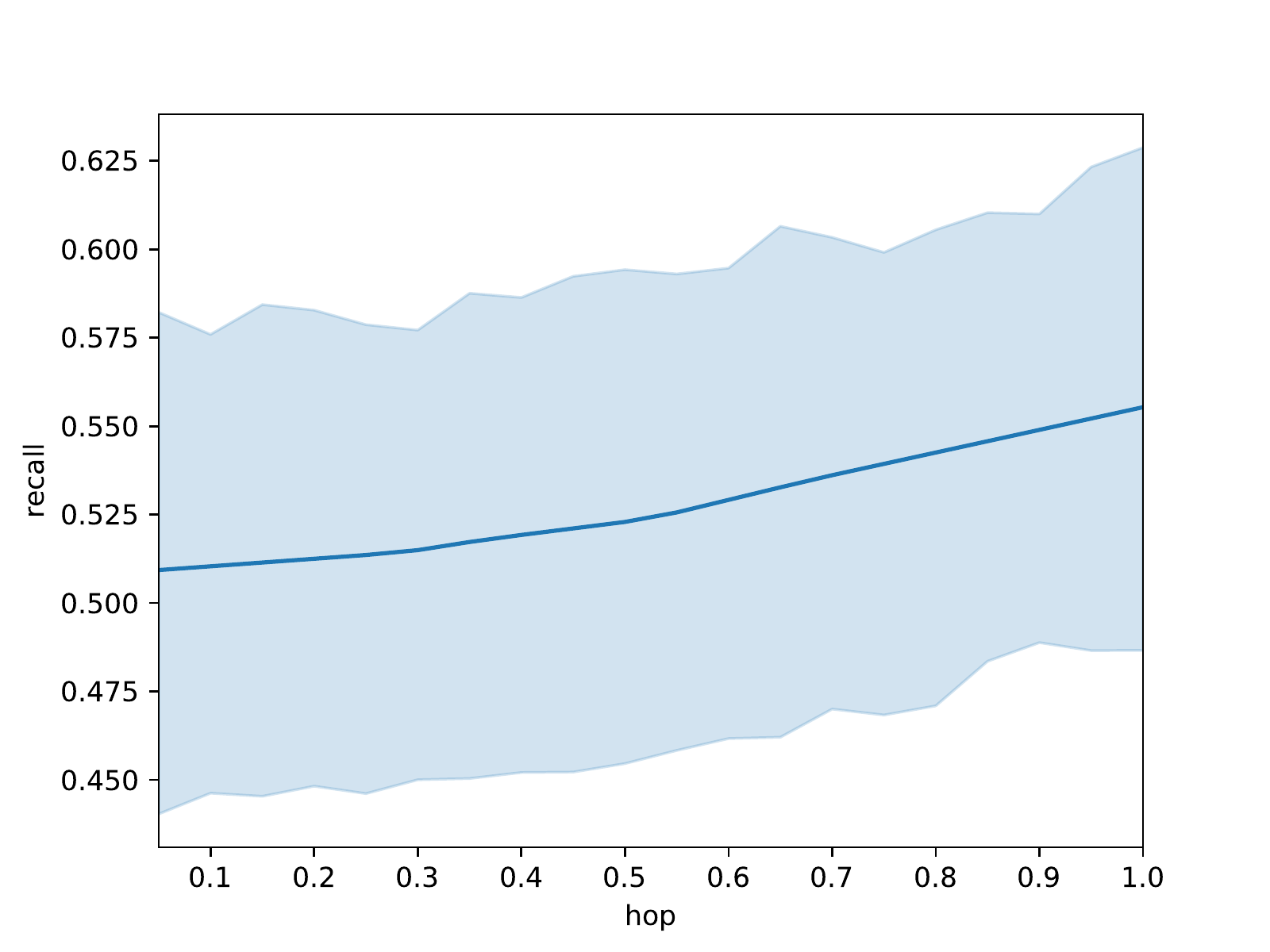}
        \caption{EPRe score}
        \label{fig:recall_score_re}        
    \end{subfigure}   
    \begin{subfigure}[t]{0.33\textwidth}
        \includegraphics[width=1.0 \textwidth]{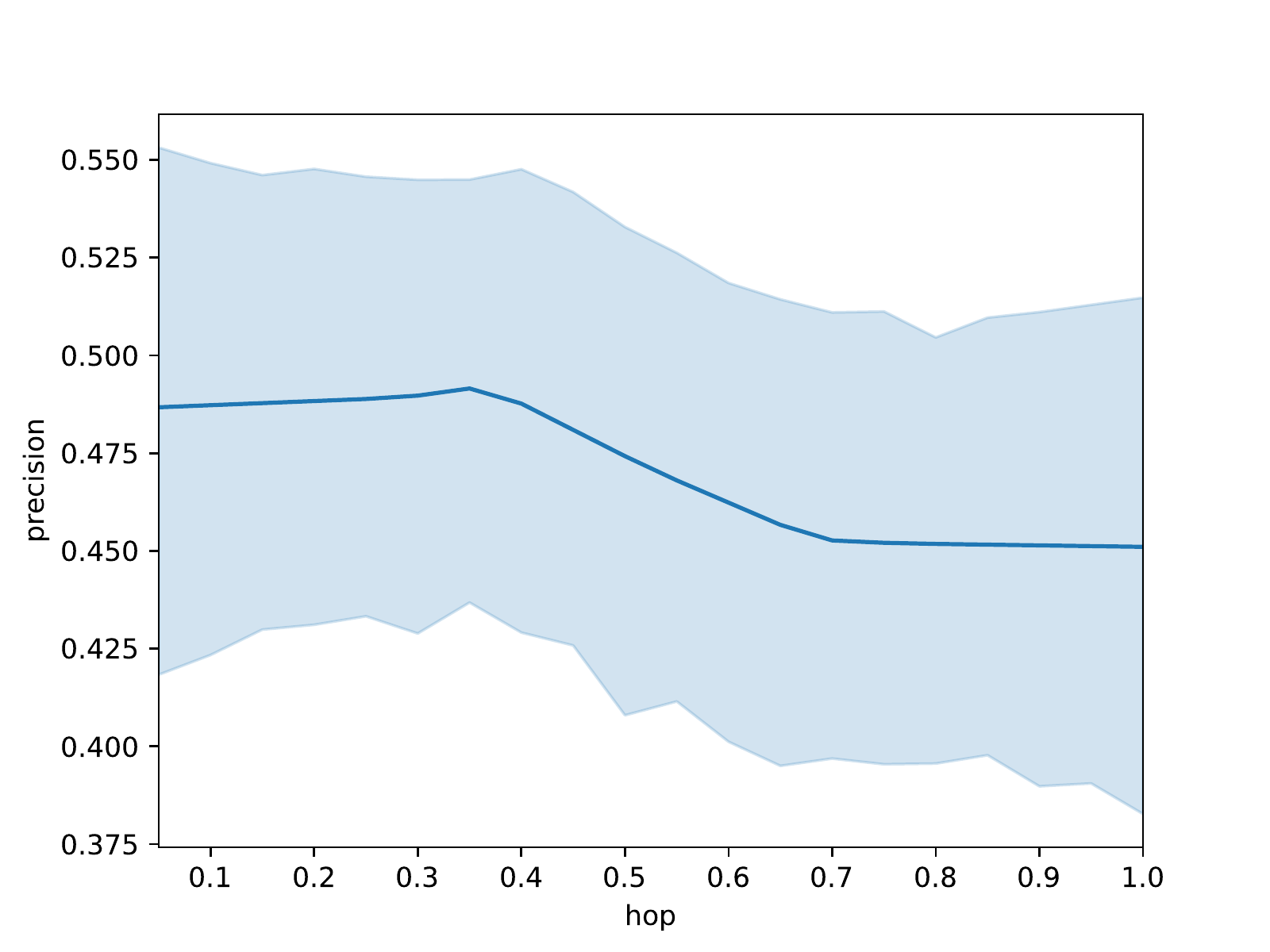}
        \caption{EPPr score}
        \label{fig:precision_score_re}
    \end{subfigure}%
    \begin{subfigure}[t]{0.33\textwidth}
        \includegraphics[width=1.0 \textwidth]{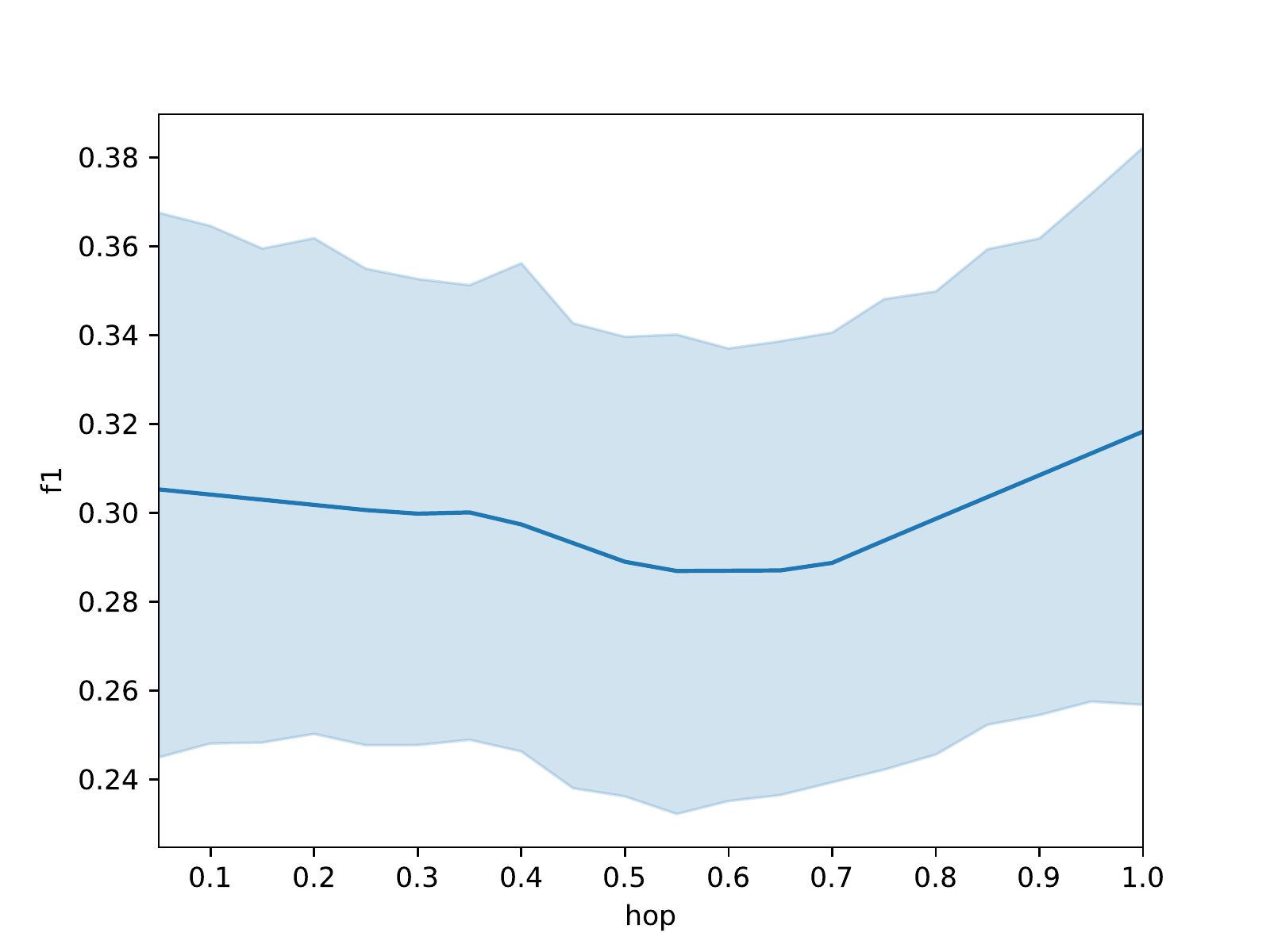}
        \caption{EPF1 score}
        \label{fig:f1_score_re}        
    \end{subfigure}  
    \\
    \begin{subfigure}[t]{0.33\textwidth}
        \includegraphics[width=1.0 \textwidth]{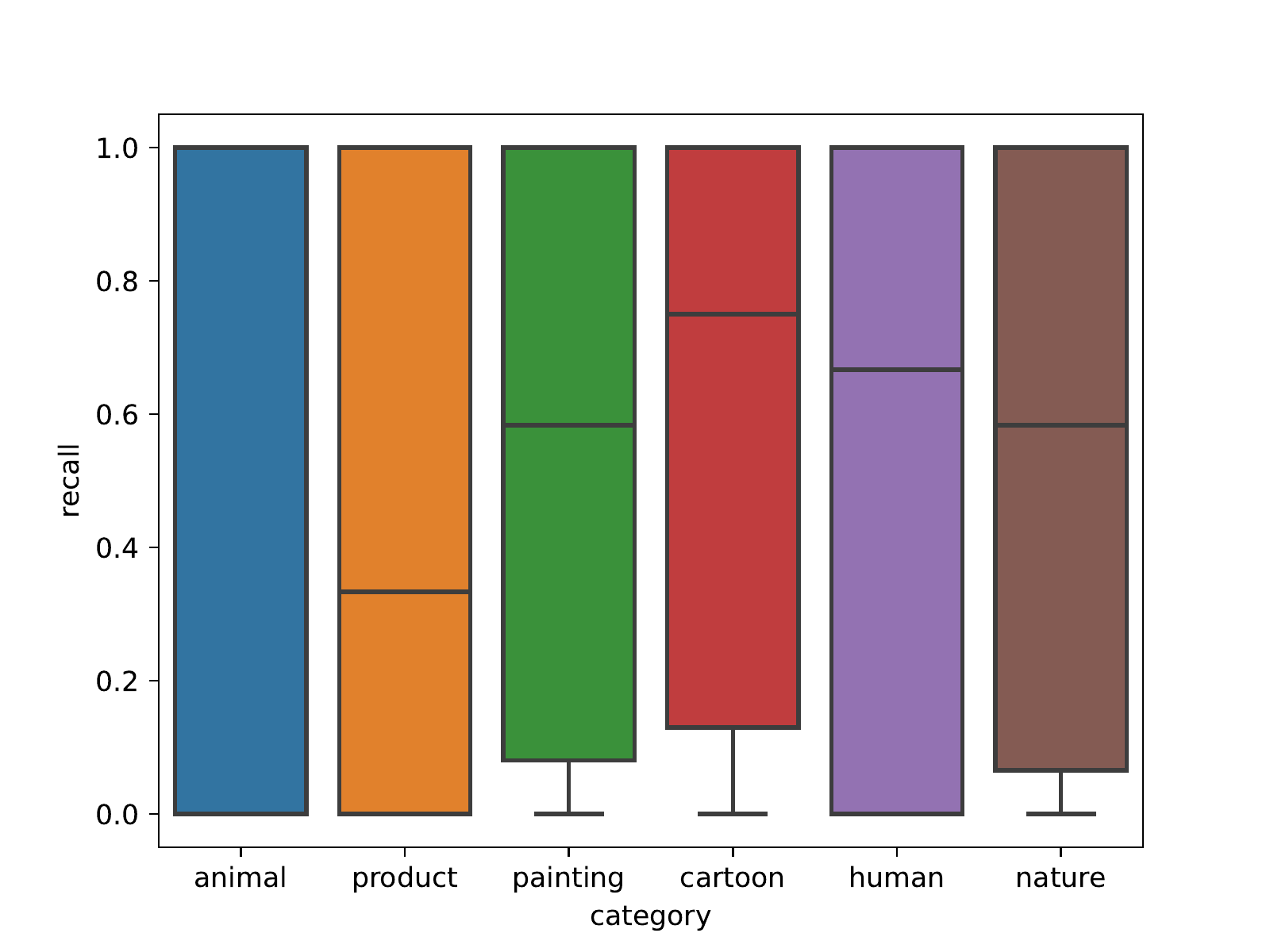}
        \caption{EPRe score}
        \label{fig:recall_score_re_ca}        
    \end{subfigure}   
    \begin{subfigure}[t]{0.33\textwidth}
        \includegraphics[width=1.0 \textwidth]{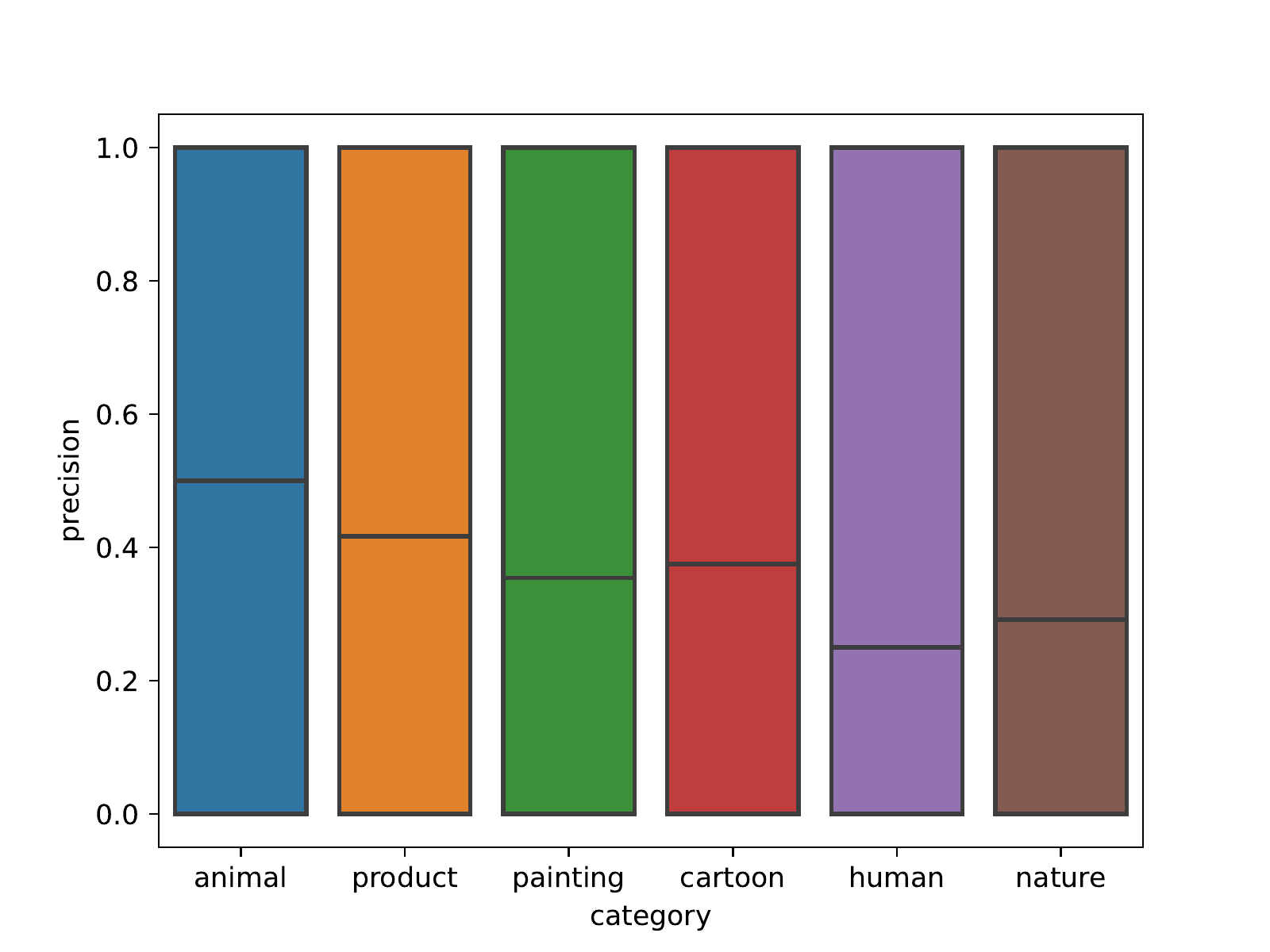}
        \caption{EPPr score }
        \label{fig:precision_score_re_ca}
    \end{subfigure}%
    \begin{subfigure}[t]{0.33\textwidth}
        \includegraphics[width=1.0 \textwidth]{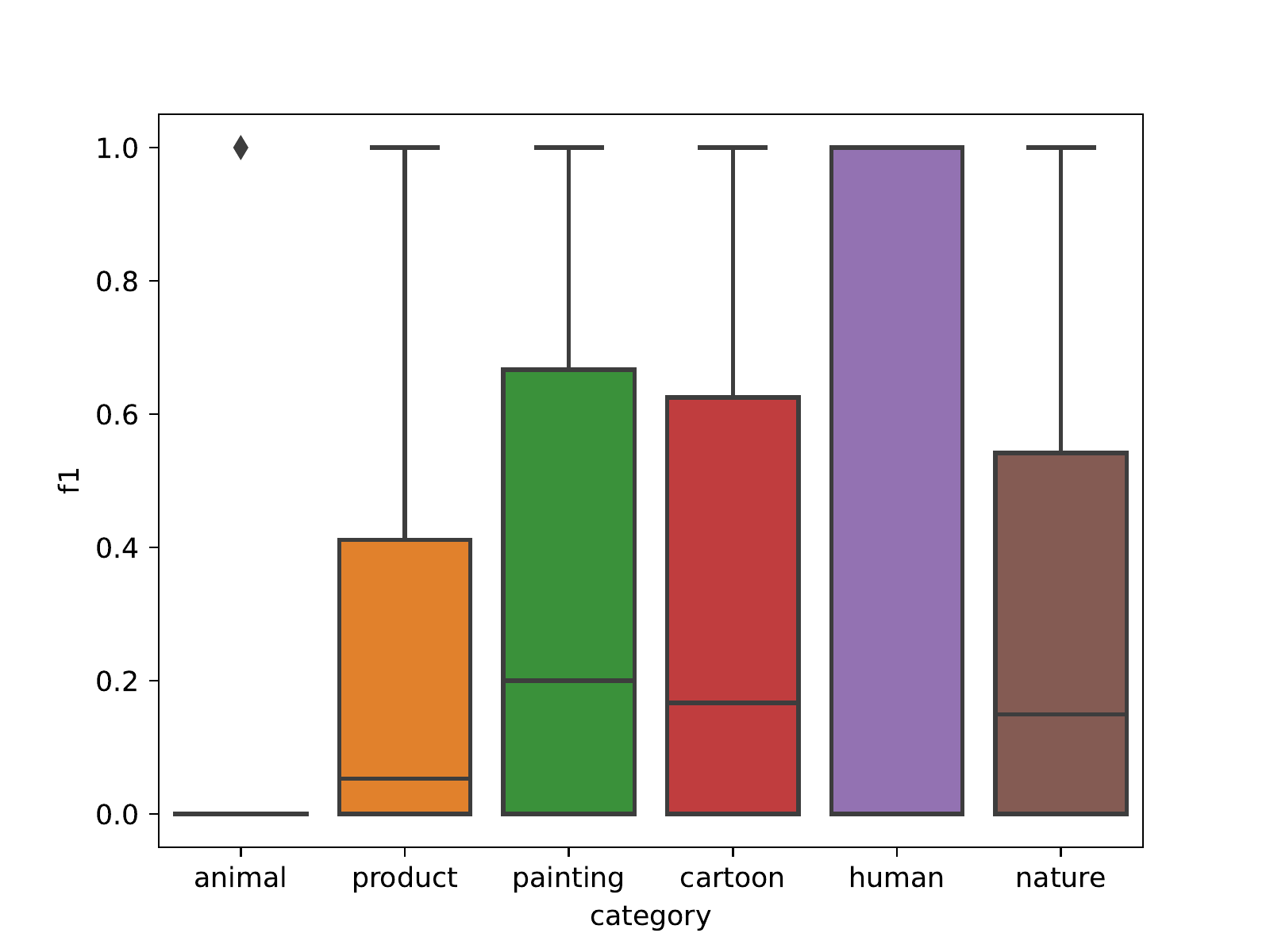}
        \caption{EPF1 score}
        \label{fig:f1_score_re_ca}        
    \end{subfigure}  
    \caption{Model performance based on Element Presence Scores: (a-c) based on hops, and (d-f) based on each image category.} %for final step.}
    \label{fig:results_element_presence_scores}
\end{figure*}

As shown in Figure~\ref{fig:recall_score_re}, the EPRe score improves as the number of hops increases in the CGVM. The EPRe score specifically evaluates the model's ability to generate images that correctly consider the presence of specific objects discussed in the conversation. In other words, as each hop progresses in the conversation, the model could add more objects to the generated images.

Figure~\ref{fig:precision_score_re} illustrates a decrease in the EPPr score as the number of hops increases in the CGVM. This observation suggests while the model has the ability to add more objects to the generated images, there is a potential for including irrelevant or incorrect objects that do not align with the ground truth or user expectations. 

Since there are conflicting trends between the EPRe and EPPr scores, the EPF1 score serves as a useful metric to provide a balanced evaluation of the model's performance.  Figure~\ref{fig:f1_score_re} demonstrates that the EPF1 score increases as the number of hops increases in the CGVM. This indicates that, overall, the model improves in terms of generating images that accurately incorporate the discussed objects while minimizing irrelevant or incorrect inclusions.

Figures~\ref{fig:recall_score_re_ca}-\ref{fig:f1_score_re_ca} illustrate the element presence scores for each image category. From the figures, it can be inferred that the animal category exhibits the highest element presence score, indicating that the generated images in this category accurately consider the presence of specific objects related to animals discussed in the conversation. This aligns with the high CLIP score observed in the animal category, suggesting a strong semantic alignment with the intended objects. Furthermore, the animal category exhibits a lower precision score. This implies that while the model incorporates more objects, some of them may not be entirely relevant or contextually appropriate based on the ground truth or user expectations. This discrepancy between EPRe and EPPr highlights the challenge of achieving a balance between including relevant objects and avoiding the addition of irrelevant or incorrect ones. Similarly, the human category shows a high EPRe score, indicating that the model successfully includes human-related objects discussed in the conversation. However, it also exhibits a lower precision score, suggesting that the generated images may still contain some irrelevant or contextually inappropriate elements. 

Results for IoU scores are presented in Table \ref{tab:iou}. The benchmark model achieves a very high score on a sample but overall for all of the scores in this category, it achieves around 0.2 and 0.3. Common-IoU score achieves a higher value and it is expected because of its nature, common objects are more likely to have similar bounding boxes and get less zero values for the objects that do not exist.

\begin{table}[]
    \centering
    \begin{tabular}{|l||l|l|l|}\hline
         \textbf{Scores} &  \textbf{Mean} & \textbf{Std} & \textbf{Maximum}\\ \hline \hline
         Common-IoU&0.298925&0.199805&0.789078\\ \hline
         Precision-IoU&0.223916&0.200435&0.789078\\ \hline
         Recall-IoU&0.273368&0.203606&0.789078\\ \hline
    \end{tabular}
    \caption{Model performance based on IoU scores.}
    \label{tab:iou}
\end{table}

\section{Discussion}
%In the upcoming years, CGVMs are anticipated to experience significant advancements, largely attributed to the utilization of LLMs and the continuous development of vision models. However, despite the progress that is expected to be made, there are several challenges that must be addressed. This section aims to highlight some of these challenges, drawing on the evidence obtained from the evaluation section.

Some key challenges we observed in the evaluation of CGVMs were: 1) the tendency of the LLMs to hallucinate and introduce additional details without fully considering the context of the conversation, 2)
 CGVM's failure to pay sufficient attention to details, and 3) ensuring that the generated images contain all significant elements discussion in the input conversation summary.

\begin{figure*}[t]
    \centering
    \begin{subfigure}[t]{0.9\textwidth}
         \includegraphics[width=1.0 \textwidth]{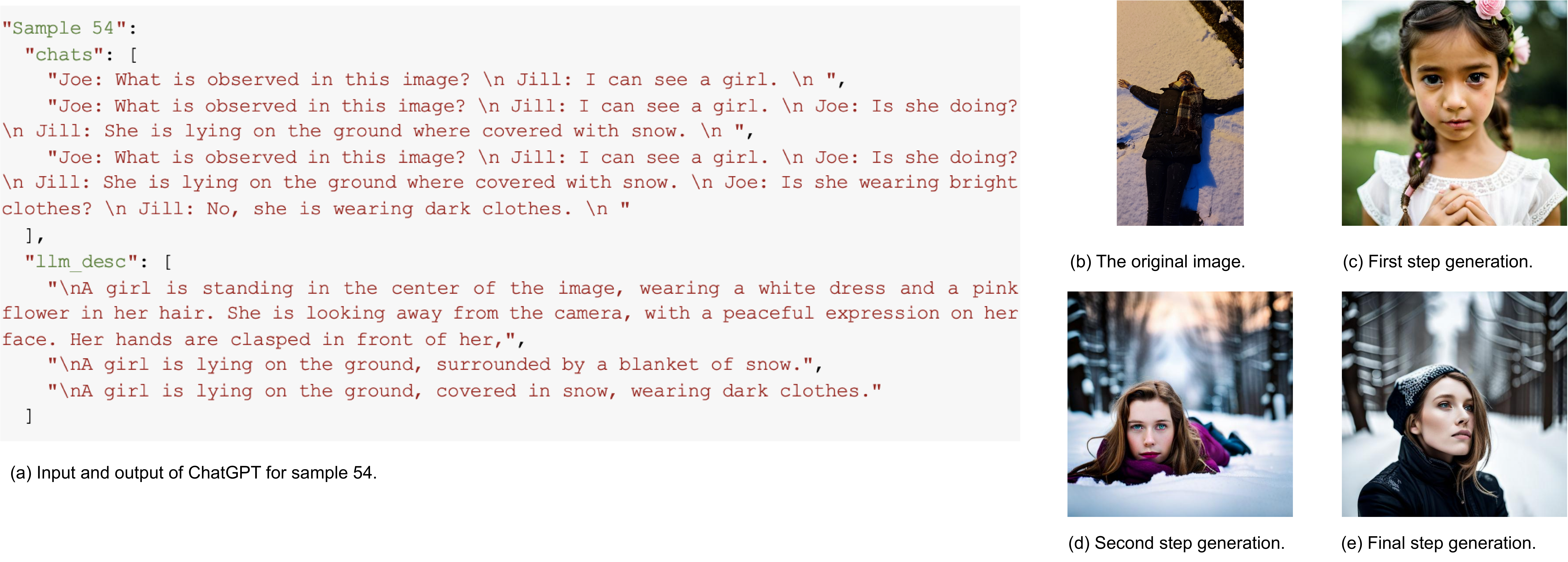}
         \caption{Sample \#54}
        \label{fig:first_evidence_sample54}      
    \end{subfigure}   \\
    \begin{subfigure}[t]{0.9\textwidth}
        \includegraphics[width=1.0 \textwidth]{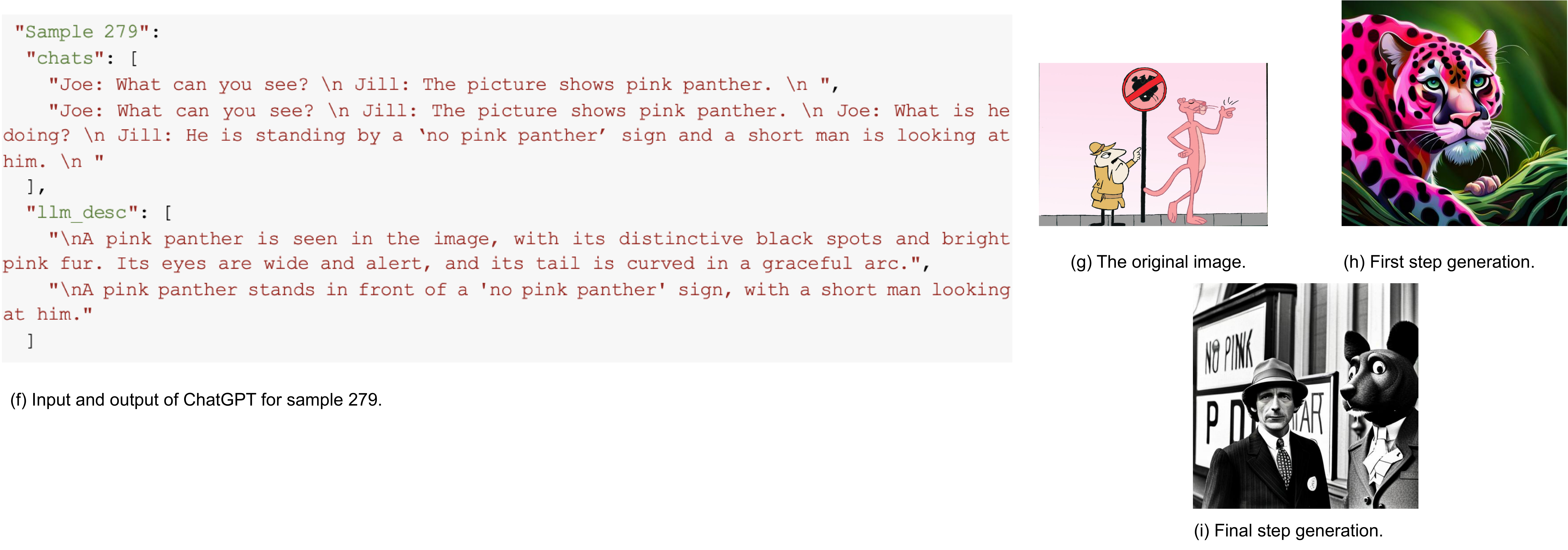}
        \caption{Sample \#279}
        \label{fig:first_evidence_sample279}
    \end{subfigure} 
    \caption{Some data samples that highlight how the CGVM model's visual component forgets important details from previous hops.}
    \label{fig:first_evidence}
\end{figure*}

One of the key challenges we observed in the evaluation of CGVMs is the tendency of the LLMs to introduce additional details without fully considering the context of the conversation. For example, Figure~\ref{fig:first_evidence_sample54} shows  sample \#$54$ where ``Jill'' answers that ``I can see a girl'' , while the LLM model adds descriptions such as "the girl is standing in the center of the image" or ``the girl is wearing a white dress and a pink flower in her hair.''  These details may not be explicitly mentioned or requested in the conversation, indicating that the LLM component tends to be proactive in generating additional information. In the subsequent hop, the LLM tends to remove the previously added details and  the vision model generates the new image based on the updated conversation context.  This is the main reason that can help explain the observed increase in the EPRe and decrease in the EPPr in CGVMs. The inclusion of more details may also introduce a higher likelihood of including irrelevant or incorrect objects, leading to a decrease in precision. 

The second challenge we encountered was, the CGVM fails to pay sufficient attention to details. For example in the final hop of sample \#$54$, as shown in Figure~\ref{fig:first_evidence_sample54}, it is noticed that the vision model neglects information about the girl lying on the snow and primarily focuses on incorporating the newly provided details about the girl's clothes.  Additionally, in another example as shown in Figure~\ref{fig:first_evidence_sample279}, the model initially generates an image of a pink panther in the first step, and the output indicates that the model does not know the pink panther as a specific cartoon figure. However, as the conversation progresses and reaches its conclusion, the final output disregards the color aspect and includes a black and white panther instead. This demonstrates how the model can neglect certain details, such as color information, throughout the course of the dialogue.

\begin{figure*}[h]
    \centering
    \begin{subfigure}[t]{0.5\textwidth}
        \includegraphics[width=1.0 \textwidth]{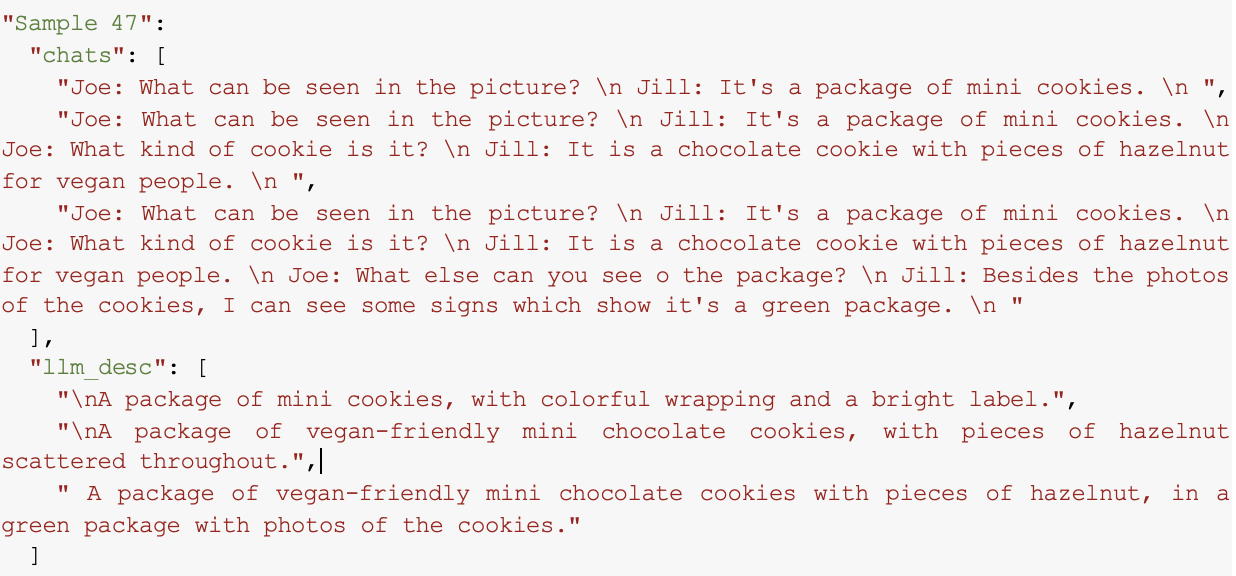}
        \caption{Input and output of ChatGPT for sample \#47}
        \label{fig:second_evidence_conv3}        
    \end{subfigure} 
    \begin{subfigure}[t]{0.2\textwidth}
        \includegraphics[width=1.0 \textwidth]{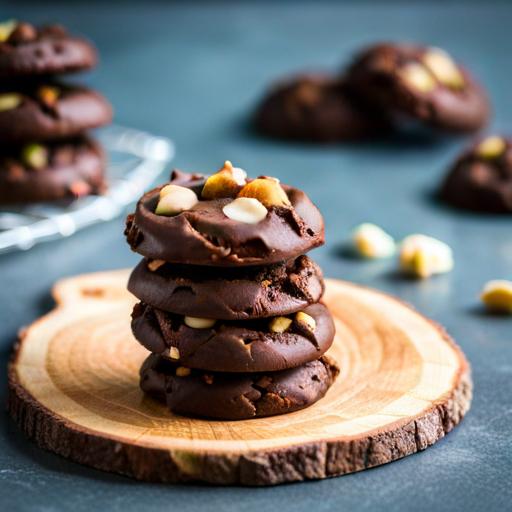}
        \caption{Output of DreamStudio for sample \#47}
        \label{fig:second_evidence_output3}        
    \end{subfigure}  
    \caption{An example that illustrates how the CGVM visual component disregarded the term ``package'' when generating the output.}
    \label{fig:second_evidence_output}
\end{figure*}

Third, it is a challenge to coordinate the LLM and vision model such that the generated images encompass all significant elements discussed in the conversation. For instance, in Figure~\ref{fig:second_evidence_conv3}, the input (original conversation) and output of ChatGPT are provided. When the term "package" is mentioned in the conversation within the product category, although the LLM can recognize the term and include it in the summary, the vision model completely disregards it, as shown in Figure~\ref{fig:second_evidence_output3}. Consequently, the output generated by the model does not include any package.  This observation highlights a challenge in maintaining consistency and comprehensiveness throughout the conversation. While the LLM may introduce new information, it is crucial for the vision model to consider and retain important details mentioned earlier in the conversation. Neglecting relevant information can result in an incomplete representation of the scene. Introducing a memory mechanism in CGVMs can address the challenge of retaining and effectively utilizing both previously mentioned details and new information. By incorporating a memory component, the model can store relevant information from previous hops and access it when generating subsequent images. 

The evaluation results shed light on the appropriate metrics for investigating the performance of CGVMs. The findings indicate that computational FR-IQA scores may not be reliable options for assessing the step-by-step improvements in generated images for each hop. These metrics are highly sensitive to even slight changes. On the other hand, semantic FR-IQA measurements, such as the CLIP score, prove to be valuable in evaluating the progression and enhancement of generated images as the number of hops increases. They provide a quantitative measure of the increasing similarity between the generated images and the original image. Additionally, the presence of specific objects is crucial, and metrics like Element Presence Scores play a critical role in assessing the generated images in terms of the presence of desired elements or objects. Semantic similarity and element-wise precision and recall are not always directly correlated. It is possible for a generated image to contain the same objects as the ground truth image, yet lack semantic similarity. Differences in background, context, or depicted actions can contribute to the semantic disparity between the two images. Conversely, two images can have a close semantic meaning while exhibiting fewer interactions on the element level. These nuances highlight the importance of employing multiple evaluation scores. By considering different metrics, we can capture various aspects of image quality and better understand the strengths and limitations of the generative model. Furthermore, it is crucial to acknowledge that metrics like CLIP score and element-wise scores do not consider the size and location of the elements in the images. For instance, if an image is mirrored, it may receive high scores in terms of CLIP and element presence. Also, if the generated objects are smaller in size compared to the ground truth objects, the same issue will occur. While IOU scores do not yield identical values in these two scenarios. IOU scores take into account both the size and location of the objects, providing a more comprehensive assessment.

We emphasize here that the primary objective of this paper was to identify suitable evaluation methodologies that can be applied to a wide range of CGVMs.  The intention was not to solely assess the performance of specific models including ChatGPT or DreamStudio. Rather, these models were utilized as means to explore solutions for evaluating CGVMs.

\section{Conclusion}
This paper introduces \textit{ConvGenVisMo}, a novel evaluation task for conversational generative vision models (CGVMs), along with the first benchmark dataset specifically designed for this task. The dataset comprises diverse image-conversation pairs across various image categories. \textit{ConvGenVisMo} also presents a comprehensive suite of evaluation metrics, combining existing and newly developed measures, to effectively assess the performance of CGVMs.  The results obtained from the evaluation emphasize the importance of employing a combination of metrics to comprehensively evaluate the performance of CGVMs.

\section*{Acknowledgement}
The authors express their gratitude to 1) Shervin Minaee for reviewing this work and providing very insightful comments, and 2) all individuals who voluntarily donated their images for this study.


\nocite{langley00}

\bibliography{example_paper}

\begin{thebibliography}{17}
\providecommand{\natexlab}[1]{#1}
\providecommand{\url}[1]{\texttt{#1}}
\expandafter\ifx\csname urlstyle\endcsname\relax
  \providecommand{\doi}[1]{doi: #1}\else
  \providecommand{\doi}{doi: \begingroup \urlstyle{rm}\Url}\fi

\bibitem[cha()]{chatgpt}
Openai. {ChatGPT}.
\newblock \url{https://chat.openai.com/}.

\bibitem[Carion et~al.(2020)Carion, Massa, Synnaeve, Usunier, Kirillov, and
  Zagoruyko]{carion2020end}
Carion, N., Massa, F., Synnaeve, G., Usunier, N., Kirillov, A., and Zagoruyko,
  S.
\newblock End-to-end object detection with transformers.
\newblock In \emph{Computer Vision--ECCV 2020: 16th European Conference,
  Glasgow, UK, August 23--28, 2020, Proceedings, Part I 16}, pp.\  213--229.
  Springer, 2020.

\bibitem[Christiano et~al.(2017)Christiano, Leike, Brown, Martic, Legg, and
  Amodei]{christiano2017deep}
Christiano, P.~F., Leike, J., Brown, T., Martic, M., Legg, S., and Amodei, D.
\newblock Deep reinforcement learning from human preferences.
\newblock \emph{Advances in neural information processing systems}, 30, 2017.

\bibitem[Deng et~al.(2009)Deng, Dong, Socher, Li, Li, and
  Fei-Fei]{deng2009imagenet}
Deng, J., Dong, W., Socher, R., Li, L.-J., Li, K., and Fei-Fei, L.
\newblock Imagenet: A large-scale hierarchical image database.
\newblock In \emph{2009 IEEE conference on computer vision and pattern
  recognition}, pp.\  248--255. Ieee, 2009.

\bibitem[et~al.(2022{\natexlab{a}})]{chowdhery2022palm}
et~al., A.~C.
\newblock Palm: Scaling language modeling with pathways, 2022{\natexlab{a}}.

\bibitem[et~al.(2022{\natexlab{b}})]{thoppilan2022lamda}
et~al., R.~T.
\newblock Lamda: Language models for dialog applications, 2022{\natexlab{b}}.

\bibitem[Everingham et~al.(2010)Everingham, Van~Gool, Williams, Winn, and
  Zisserman]{Everingham10}
Everingham, M., Van~Gool, L., Williams, C. K.~I., Winn, J., and Zisserman, A.
\newblock The pascal visual object classes (voc) challenge.
\newblock \emph{International Journal of Computer Vision}, 88\penalty0
  (2):\penalty0 303--338, June 2010.

\bibitem[Lin et~al.(2014)Lin, Maire, Belongie, Hays, Perona, Ramanan,
  Doll{\'a}r, and Zitnick]{lin2014microsoft}
Lin, T.-Y., Maire, M., Belongie, S., Hays, J., Perona, P., Ramanan, D.,
  Doll{\'a}r, P., and Zitnick, C.~L.
\newblock Microsoft coco: Common objects in context.
\newblock In \emph{Computer Vision--ECCV 2014: 13th European Conference,
  Zurich, Switzerland, September 6-12, 2014, Proceedings, Part V 13}, pp.\
  740--755. Springer, 2014.

\bibitem[Mittal et~al.(2012)Mittal, Moorthy, and Bovik]{mittal2012no}
Mittal, A., Moorthy, A.~K., and Bovik, A.~C.
\newblock No-reference image quality assessment in the spatial domain.
\newblock \emph{IEEE Transactions on image processing}, 21\penalty0
  (12):\penalty0 4695--4708, 2012.

\bibitem[OpenAI(2023)]{openai2023gpt4}
OpenAI.
\newblock Gpt-4 technical report, 2023.

\bibitem[Radford et~al.(2021)Radford, Kim, Hallacy, Ramesh, Goh, Agarwal,
  Sastry, Askell, Mishkin, Clark, Krueger, and
  Sutskever]{DBLP:journals/corr/abs-2103-00020}
Radford, A., Kim, J.~W., Hallacy, C., Ramesh, A., Goh, G., Agarwal, S., Sastry,
  G., Askell, A., Mishkin, P., Clark, J., Krueger, G., and Sutskever, I.
\newblock Learning transferable visual models from natural language
  supervision.
\newblock \emph{CoRR}, abs/2103.00020, 2021.
\newblock URL \url{https://arxiv.org/abs/2103.00020}.

\bibitem[Ramesh et~al.(2021)Ramesh, Pavlov, Goh, Gray, Voss, Radford, Chen, and
  Sutskever]{DBLP:journals/corr/abs-2102-12092}
Ramesh, A., Pavlov, M., Goh, G., Gray, S., Voss, C., Radford, A., Chen, M., and
  Sutskever, I.
\newblock Zero-shot text-to-image generation.
\newblock \emph{CoRR}, abs/2102.12092, 2021.
\newblock URL \url{https://arxiv.org/abs/2102.12092}.

\bibitem[Rombach et~al.(2021)Rombach, Blattmann, Lorenz, Esser, and
  Ommer]{DBLP:journals/corr/abs-2112-10752}
Rombach, R., Blattmann, A., Lorenz, D., Esser, P., and Ommer, B.
\newblock High-resolution image synthesis with latent diffusion models.
\newblock \emph{CoRR}, abs/2112.10752, 2021.
\newblock URL \url{https://arxiv.org/abs/2112.10752}.

\bibitem[Touvron et~al.(2023)Touvron, Lavril, Izacard, Martinet, Lachaux,
  Lacroix, Rozi{\`e}re, Goyal, Hambro, Azhar, et~al.]{touvron2023llama}
Touvron, H., Lavril, T., Izacard, G., Martinet, X., Lachaux, M.-A., Lacroix,
  T., Rozi{\`e}re, B., Goyal, N., Hambro, E., Azhar, F., et~al.
\newblock Llama: Open and efficient foundation language models.
\newblock \emph{arXiv preprint arXiv:2302.13971}, 2023.

\bibitem[Wang \& Bovik(2002)Wang and Bovik]{wang2002universal}
Wang, Z. and Bovik, A.~C.
\newblock A universal image quality index.
\newblock \emph{IEEE signal processing letters}, 9\penalty0 (3):\penalty0
  81--84, 2002.

\bibitem[Wang et~al.(2004)Wang, Bovik, Sheikh, and Simoncelli]{1284395}
Wang, Z., Bovik, A., Sheikh, H., and Simoncelli, E.
\newblock Image quality assessment: from error visibility to structural
  similarity.
\newblock \emph{IEEE Transactions on Image Processing}, 13\penalty0
  (4):\penalty0 600--612, 2004.
\newblock \doi{10.1109/TIP.2003.819861}.

\bibitem[Wu et~al.(2023)Wu, Yin, Qi, Wang, Tang, and Duan]{wu2023visual}
Wu, C., Yin, S., Qi, W., Wang, X., Tang, Z., and Duan, N.
\newblock Visual chatgpt: Talking, drawing and editing with visual foundation
  models.
\newblock \emph{arXiv preprint arXiv:2303.04671}, 2023.

\end{thebibliography}
\bibliographystyle{icml2023}

%%%%%%%%%%%%%%%%%%%%%%%%%%%%%%%%%%%%%%%%%%%%%%%%%%%%%%%%%%%%%%%%%%%%%%%%%%%%%%%
%%%%%%%%%%%%%%%%%%%%%%%%%%%%%%%%%%%%%%%%%%%%%%%%%%%%%%%%%%%%%%%%%%%%%%%%%%%%%%%
% APPENDIX
%%%%%%%%%%%%%%%%%%%%%%%%%%%%%%%%%%%%%%%%%%%%%%%%%%%%%%%%%%%%%%%%%%%%%%%%%%%%%%%
%%%%%%%%%%%%%%%%%%%%%%%%%%%%%%%%%%%%%%%%%%%%%%%%%%%%%%%%%%%%%%%%%%%%%%%%%%%%%%%
\newpage
\onecolumn

%\newpage

\end{document}